\documentclass{article}

\pdfoutput=1

\usepackage{spconf,amsmath,epsfig}
\usepackage{graphicx}
\usepackage{epstopdf}
\usepackage{amssymb}
\usepackage{algorithm}
\usepackage[noend]{algpseudocode}
\usepackage{caption}
\usepackage{subcaption}
\usepackage{multirow}
\usepackage{color}

\newcommand{\figwidth}{6.5}
\newcommand{\figheight}{8}

\begin{document}\sloppy

\def\x{{\mathbf x}}
\def\L{{\cal L}}

\title{Learning a Pose Lexicon for Semantic Action Recognition}
%
\name{Lijuan Zhou, Wanqing Li, Philip Ogunbona}
\address{School of Computing and Information Technology \\
University of Wollongong, Keiraville, NSW 2522, Australia\\
lz683@uowmail.edu.au, wanqing@uow.edu.au, philipo@uow.edu.au}
%
%
%

\maketitle

\begin{abstract}
This paper presents a novel method for learning a pose lexicon 
comprising semantic poses defined by textual instructions and 
their associated visual poses defined by visual features. 
The proposed method simultaneously takes two input streams, 
semantic poses and visual pose candidates, and statistically learns a mapping between them to construct the lexicon.  With the learned lexicon, action 
recognition can be cast as the problem of finding the maximum translation 
probability of a sequence of semantic poses given a stream of visual 
pose candidates. Experiments evaluating pre-trained and zero-shot action
recognition conducted on MSRC-12 gesture and WorkoutSu-10 exercise datasets 
were used to verify the efficacy of the proposed method.
\end{abstract}
%

\section{Introduction}
\label{i}
Human action recognition is currently one of the most active research topics in 
multimedia content analysis. Most recognition models are typically constructed 
from low to middle level visual spatio-temporal features and directly 
associated with class labels~\cite{li2008expandable,weinland2011survey,wang2014mining,wang2015convnets,wang2015action}. In particular, many 
methods~\cite{li2008expandable,li2010action,wang2013approach,eweiwi2014efficient} have been developed 
based on the concept that an action can be well represented by a sequence of key 
or salient poses and these salient poses can be identified through visual 
features alone. However, these salient poses often do not necessarily possess 
semantic significance thus leading to the so-called semantic gap. We refer to 
the salient poses defined using visual features as ``visual poses".

The textual instruction of how a simple or elementary action should be 
performed is often expressed using Talmy's typology of 
motion~\cite{Talmy2003}. In this typology a motion event is one 
in which an entity experiences a change in location with 
respect to another object and such event can be described by 
four semantic elements including \textit{Motion (M)}, \textit{Figure (F)}, 
\textit{Ground (G)} and \textit{Path (P)}, where \textit{M} 
refers to the movement of an entity \textit{F} that changes its 
location relative to the reference object \textit{G} along the trajectory 
\textit{P} (including a start and an end). 
For a complex action or an action that involves many body parts, a sequence 
of basic textual instructions will suffice. Each instruction, describing a part 
of the action, will be made up of the four elements. A simplification of the 
trajectory \textit{P} for human actions entails retaining the starting $P_s$ 
and end $P_e$, status or configuration of the body parts, and ignoring 
intermediate parts of the trajectory. We refer to both $P_s$ and $P_e$ 
as ``semantic poses". 
Hence, an action can be described by a sequence of semantic poses if 
\textit{F} is defined as the whole human body. Alternatively, if 
\textit{F} refers to a body part, multiple sequences of semantic poses 
can be used. Semantic poses can be obtained by parsing the textual instructions.

This paper proposes a method to construct a pose lexicon comprising a set of 
semantic poses and the corresponding visual poses, by learning a 
mapping between them. It is assumed that for each action there is a textual 
instruction from which a set of semantic poses can be extracted through natural 
language parsing~\cite{petrov2006learning} and that for each action sample a sequence of visual pose candidates can 
be extracted. The mapping task is formulated as a problem of machine 
translation. With the learned lexicon, action recognition can be considered as a 
problem of finding the maximum posterior probability of a given sequence of visual 
pose candidates being generated from a given sequence of semantic poses. This is 
equivalent to 
determining how likely the given sequence of visual pose candidates follow a sequence of 
semantic poses. Such a lexicon bridges the gap between the semantics and visual 
features and offers a number of advantages including text-based action retrieval 
and summarization, recognition of actions with small or even zero training 
samples (also known as zero-short recognition), and easy growth of semantic 
poses for new action recognition since poses in the lexicon are sharable by many actions.

The rest of this paper is organized as follows. Section~\ref{rw} provides a 
review of previous work related to semantic action recognition. 
The proposed method for learning pose 
lexicon and action classification is developed and formulated in  
Section~\ref{pm}. In 
Section~\ref{er}, experiments are presented to demonstrate the effectiveness of 
the proposed method in recognition tasks using MSRC-12 Kinect 
gesture~\cite{Fothergill2012}, WorkoutSU-10 exercise~\cite{Negin2013} datasets 
and novel actions extracted from the two datasets. 
Finally, the paper is concluded with remarks in Section~\ref{c}.
\section{Related work}
\label{rw}
Despite the good progress made in action recognition over the past decade, few 
studies have reported methods based on semantic learning. 
Earlier methods bridged the semantic gap using mid-level features (eg. 
visual keywords)~\cite{laptev2008learning}) obtained by quantizing low-level 
spatio-temporal features which form visual vocabulary.  However, mid-level 
features are not sufficiently robust to obtain good performance on relatively 
large action dataset. This problem has been addressed by proposing high-level 
latent semantic features to represent semantically similar mid-level features. 
Unsupervised methods~\cite{Niebles2008,Wang2009} were previously applied for 
learning latent semantics based on topic models; example include probabilistic 
latent semantic analysis~\cite{Hofmann2001} and latent Dirichlet allocation 
(LDA)~\cite{Blei2003}.
Recently, multiple layers model~\cite{yang2014hierarchical} based on LDA was proposed 
for learning local and global action semantics. The intuitive basis of using 
mid- and high-level latent semantic features is that frequently co-occurring 
low-level features are correlated at some conceptual level. It is 
noteworthy that these two kinds of semantic features have no explicit 
semantic relationship to the problem; a situation different from the proposed 
semantic poses. 

Apart from learning latent semantics of actions, other approaches focused 
on defining semantic concepts to describe action or activity related 
properties. Actions were described by a set of attributes that possess spatial 
characteristics. Unfortunately, the attributes are not specific enough 
to allow subjects to recreate the actions~\cite{Liu2011, zhang2013attribute}.  
It is also difficult to describe them as there is no a common principle for 
describing different actions. In our work, textual instructions use a common 
principle (four semantic elements) for action representation and they also 
provide unambiguous guideline for performing actions. Activity was 
represented by basic actions and corresponding participants such as subjects, 
objects and tools~\cite{rohrbach2012script, guadarrama2013youtube2text}. 
The method for representing activities ties the object and action together. The 
work presented in this paper focuses on single actions which do not depend on other objects. 

\vspace{-0.6cm}
\section{Proposed method}
\label{pm}

\begin{figure}
\begin{center}
\includegraphics[width=0.98\linewidth]{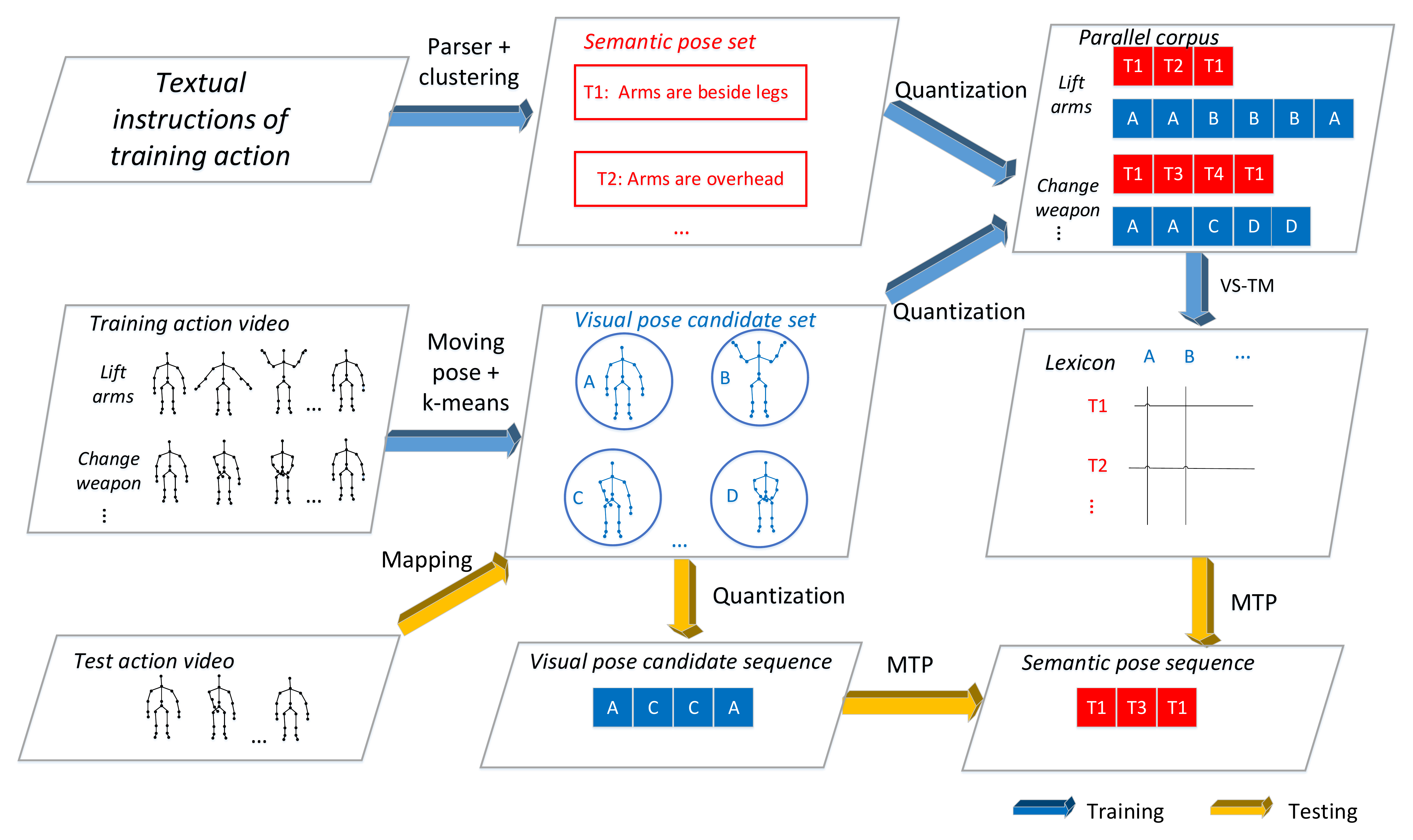}
\end{center}
\vspace{-3 mm}
   \caption{Framework of the proposed method.}
\label{fig2_frame}
\end{figure}

Visual poses can be extracted from either RGB, depth maps or skeleton 
data. In this work, we take skeleton data as an example to illustrate the 
proposed method. Inspired by the translation model for image 
annotation~\cite{Duygulu2002}, a translation model from visual poses to semantic poses, 
namely ``visual pose-to-semantic pose translation model'' (VS-TM), is 
proposed for learning a pose lexicon from skeleton data with instructions. 
Figure~\ref{fig2_frame} 
illustrates the action recognition framework based on the VS-TM. In the 
training phase, a pose lexicon is constructed in two main steps. First, a 
parallel corpus is constructed based on a stream of semantic poses and a stream 
of visual pose candidates. Second, a mapping between the two streams is learned from the 
parallel corpus to generate the lexicon for inferring optimal visual poses from the candidates. 
In the test phase, actions are classified according to the maximum 
translation probability (MTP) of a semantic pose sequence given a sequence of visual pose candidates. 

\vspace{-3 mm}
\subsection{Parallel corpus construction}
\label{pcc}
An action instance is represented by two streams; semantic pose and 
visual pose candidate. 
The parallel corpus consists of multiple such data streams that have been 
constructed from each action instance through vector quantization on the sets of semantic pose and visual pose candidate. 
In the following Sections~\ref{text_key} and~\ref{ape} we focus on how to 
construct the two sets.
\vspace{-0.3cm}
\subsubsection{Semantic poses generation}
\label{text_key}
Semantic poses are constructed based on start and end 
semantic poses $P_s$ and $P_e$. $P_s$ refers to the configuration of human body in which the 
action starts and $P_e$ indicates the point at which body parts reach a salient configuration, e.g. maximum 
extension. $P_s$ and $P_e$ are normally encoded by preposition phrases in the 
textual instruction. 
Constituency-based parser (i.e. Berkeley~\cite{petrov2006learning}) can be used 
for extracting $P_s$ and $P_e$. Note that parsing is not the focus of paper. Suppose 
an action instance contains $G$ elementary or simple actions. The semantic 
pose sequence can be written as $\{P_{s_1}, P_{e_1},\ldots, P_{s_i}, 
P_{e_i},\ldots, P_{s_G}, P_{e_G}\}$, where $P_{s_i}$ and $P_{e_i}$ denote 
semantic poses of the $i$-th elementary action.  
Once the semantic poses are extracted, similar semantic poses are replaced 
by a single symbol and the semantic pose set is easily constructed.

\vspace{-0.2cm}
\subsubsection{Visual pose candidates generation}
\label{ape}
Key frames are firstly extracted from each action instance to visually represent the start and end semantic poses. Visual pose candidates are further generated through clustering key frames of all action instance. In this paper, we consider key frames as frames when body parts reach maximum or minimum extension. If we consider skeleton joints as observations, the covariance matrix of joint positions at an instant time captures the stretch of body parts or distribution of joints in space at this instant time. Hence, the covariance of joint positions is applied for extracting key frames. 

Given an action instance containing $F$ frames, $F$ feature vectors are generated to represent this instance using a moving pose 
descriptor~\cite{Zanfir2013}. A covariance is calculated from each feature vector, resulting in a $3\times 3$ matrix. Suppose $\Sigma^f$ denote the
covariance matrix at frame $f$ $(f\in\{1,\ldots,F\})$. 
The relationships amongst the joints of the pose can be 
analysed by performing eigen-decomposition on $\Sigma^f$. For each frame we 
select the largest eigenvalue, denoted by  $\lambda^f$ and thus, produce the 
sequence $\Lambda = \{\lambda^1,\lambda^2,\ldots,\lambda^f,\ldots,\lambda^F \}$.

To reduce the impact of noise in skeletons, we smoothen the sequence $\Lambda$ along the 
time dimension, with a moving Gaussian filter of window size 5 
frames. Then frames, at whose smoothed largest eigenvalues are bigger or smaller than those of neighbouring frames, are extracted as key frames. In particular, frame $f$ 
is a key frame if the following conditions are met:
\begin{equation}
\begin{array}{l}
\lambda^f> \lambda^{f+1}\\
\lambda^f> \lambda^{f-1} 
\end{array} or \begin{array}{l}
\lambda^f< \lambda^{f+1}\\
\lambda^f< \lambda^{f-1} .
\end{array}
\label{equ2}
\end{equation}

The extracted key frames of all action instances are clustered using $k$-means algorithm and cluster centers are considered as visual pose candidates. The selection of $k$ depends on the size of semantic pose set and the 
size of semantic pose set is determined by the number of elementary actions. 
Hence, it is available before learning the lexicon. One semantic pose can be 
mapped to multiple visual pose candidates when these visual pose candidates are similar. However, 
one visual pose candidate corresponds to at most one semantic pose. Therefore, $k$ is chosen as 
equal or larger than the number of semantic poses so that any semantic pose can 
correspond to at least one visual pose candidate. 
\vspace{-4 mm}
\subsection{Lexicon Learning}
\label{apt}
\subsubsection{Formulation}
\label{lex}
Given the parallel corpus which has underlying correspondence between sequences of semantic pose and 
visual pose candidate, the problem of learning an action lexicon entails 
determining precise correspondences among the elements of the two sequences. 
Let the set of visual pose candidates be denoted by $S=\{S_1,S_2,\ldots,S_p,\ldots\}$ and 
the semantic pose set by $T=\{T_1,T_2,\ldots,T_q,\ldots\}$. The task of 
lexicon construction is converted to finding the most likely visual pose candidate given a
semantic pose, based on conditional probability $P(S_p|T_q)$. 

The underlying correspondence between the two sequences provides an opportunity to 
use machine translation to model the problem. The sequence pair encodes the 
starting and end positions of actions which are discrete units. These discrete 
units are actually analogous to words in translation model. This observation 
makes the particular word-based machine translation framework useful to our 
problem. The sequence of visual pose candidate is analogous to the source language and 
semantic pose sequence is similar to the target language. According to the 
standard word-based 
translation framework~\cite{koehn2009statistical}, learning a lexicon is 
converted to the particular problem - translation model. Hence, we develop a 
translation model from visual poses to semantic poses (VS-TM) based on the 
parallel corpus to learn a pose lexicon. 

We now illustrate the translation  model (VS-TM). Let $M_n$ denote the number 
of visual pose candidates in the $n$-th action instance. The 
sequence of visual pose candidate, after quantization, can be written as 
$s_n=\{s_{n1},s_{n2},\ldots, s_{nj},\ldots,s_{nM_{n}}\}(s_{nj}\in S).$
Similarly, if $L_n$ represents the number of semantic poses in the $n$-th 
action instance, then the semantic pose sequence of the action instance 
can be written as 
$t_n=\{t_{n1},t_{n2},\ldots,t_{ni},\ldots,t_{nL_{n}}\}(t_{ni}\in T)$. VS-TM 
finds the most likely sequence of visual pose candidate for each semantic pose sequence 
through the conditional probability $P(s_n|t_n)$.

In the word-based translation model, the conditional probability of two 
sequences is converted to the conditional probability of elements of the 
sequences. However, we 
do not know the correspondence between individual elements of the sequence 
pair. If we introduce a hidden variable $a_n$ which determines the alignment of 
$s_n$ and $t_n$, the alignment of $N$ sequence pairs form a set which can be 
written as $a=\{a_1, a_2,\ldots, a_n,\ldots, a_N \}$. Based on $a_n$, the 
translation of sequence pair is accomplished through summing conditional 
probabilities of all possible alignments. Hence, we learn VS-TM through 
element-to-element alignment models. 
$P(s_n|t_n)$ is calculated using
\begin{equation}
P(s_n|t_n)=\sum_{a_n} P(s_n, a_n|t_n).
\label{equ4}
\end{equation}

If each visual pose candidate in the sequence can be aligned to at most one semantic 
pose, we guarantee that a visual pose candidate corresponds to only one semantic 
pose. To ensure this constraint, we construct the alignment from visual to 
semantic pose sentence. The alignment of the $n$-th instance can be written 
as $a_n=\{a_{n1}, a_{n2}, \ldots, a_{nj}, \ldots, a_{nM_{n}}\}(a_{nj}\in[0, 
L_n])$, where $a_{nj}$ represents the alignment position of the $j$-th 
visual pose candidate. If the $j$-th visual pose candidate is aligned to the $i$-th 
semantic pose, we write, $a_{nj}=i$. $a_{nj}=0$ refers to the situation in 
which no semantic pose corresponds to this visual pose candidate;  this happens when 
visual pose candidate is noisy. 
According to alignment $a_n$, Equation~\eqref{equ4} can be extended through 
structuring $P(s_n, a_n|t_n)$ without loss of generality by chain rule as 
follows:
\begin{equation}
\begin{split}
P(s_n|t_n)&=\sum_{a_n}\prod_{j=1}^{M_n} P(a_{nj}|s_{n1}^{n(j-1)},a_{n1}^{n(j-1)},t_{n1}^{nL_n})\\
& \times P(s_{nj}|s_{n1}^{n(j-1)},a_{n1}^{nj},t_{n1}^{nL_n}),
\end{split}
\label{equ5}
\end{equation}
where the first item determines alignment probability, 
the second encodes translation probability and 
$x_{n1}^{nj}=\{x_{n1},x_{n2},\ldots,x_{nj}\}(x\in (s,t,a))$. Since 
lexicon acquisition aims to find conditional probability among visual 
and semantic poses, we further assume that alignment probabilities are equal 
(i.e. $\frac{1}{L_n+1}$) and $s_{nj}$ depends only on the sequence element at 
$a_{nj}$ position which is $t_{na_{nj}}$ (equal to $t_{ni}$). Hence, Equation~\eqref{equ5} can be 
rewritten as 
\begin{equation}
P(s_n|t_n)=\sum_{a_{nj}=0}^{L_n}\prod_{j=1}^{M_n} \frac{1}{L_n+1}P(s_{nj}|t_{n{a_{nj}}}),
\label{equ6}
\end{equation}
where the translation probability is constrained through $\sum_{S_p}P(S_p|T_q)=1$ for any $T_q$.

For $N$ action instances in the training parallel corpus, the proposed model 
VS-TM aims to maximize the translation probability $P(s|t)$ through
\begin{equation}
P(s|t)=\prod_{n=1}^N \prod_{j=1}^{M_n} \sum_{i=0}^{L_n} \frac{1}{L_n+1}P(s_{nj}|t_{ni}).
\label{equ3}
\end{equation}
Here, it is easy to verify that the sum can be interchanged in Equation~\eqref{equ6}.
\vspace{-3 mm}
\subsubsection{Optimization}
The expectation maximization (EM) algorithm is invoked for the optimization by 
mapping the translation probability $P(S_p|T_q)$ to parameter $\theta$ 
and alignment $a$ to the unobserved data. 
The likelihood function $\mathcal{L}$ is defined as
\begin{equation}
\mathcal{L}_{\theta}(s,t,a)=\prod_{n=1}^N\prod_{j=1}^{M_n} \sum_{i=0}^{L_n} 
\frac{1}{L_n+1}P(s_{nj}|t_{ni}).
\end{equation}
Maximizing likelihood function $\mathcal{L}$ is further extended to seek 
an unconstrained extremum of auxiliary function
\begin{equation}
\begin{split}
h(P, \beta)\equiv&\prod_{n=1}^N\prod_{j=1}^{M_n} \sum_{i=0}^{L_n} \frac{1}{L_n+1}P(s_{nj}|t_{ni})\\
&-\sum_{T_q}\beta(\sum_{S_p}P(S_p|T_q)-1)
\end{split}
\end{equation}

In the E-step, the posterior probability among alignment is calculated by
\begin{equation}
P_{\theta}(a_{nj}|s_{nj}, t_{ni})=\frac{P(s_{nj}|t_{ni})}{\sum_{i=0}^{L_n} P(s_{nj}|t_{ni})}
\end{equation}
In the M-step, parameter $\theta$ is updated through
\begin{equation}
\begin{split}
P(S_p|T_q)& = \beta^{-1}\sum_{n=1}^N\sum_{j=1}^{M_n} 
\sum_{i=0}^{L_n}P_{\theta}(a_{nj}|s_{nj}, 
t_{ni})\\
 & \times \delta(S_p,s_{nj})\delta(T_q,t_{ni}).
\end{split}
\end{equation}
Here, $\delta(.,.)$ is 1 if two elements are equal and 0 otherwise. $\beta$ 
normalizes the probabilities.
\vspace{-3 mm}
\subsection{Action classification}
\label{class}
Once the translation model from visual pose candidates to semantic poses is learned, 
the task of action classification is converted to finding the most likely 
semantic pose sequence given a sequence of visual pose candidate. This is the decoding 
process in machine translation system~\cite{koehn2009statistical}. 
Since textual instructions of all action classes are available, we reduce 
search space to the possible solution space containing 
instructions of all trained actions. 

Let $s^{test}=\{s_{1}',\ldots,s_{j}',\ldots, s_{m}'\}(s_{j}'\in S)$ denote 
the sequence of visual pose candidate in a test action instance and $t^{test}=\{t_1',\ldots, 
t_i',\ldots, t_l'\}(t_i'\in T)$, its semantic pose sequence. The 
alignment $a^{test}$ is written as $a^{test}=\{a_1', \ldots, a_j', \ldots, a_m'\}$. 
Given the model parameter $\theta$, action is classified based on 
$P(s^{test}|t^{test})$ which is calculated through finding the best alignment to 
avoid summing all possible alignment probability. In particular, it is 
formulated as
\begin{equation}
\{t^{test}, a^{test}\}=\operatorname*{arg\,max}_{\{t^{test}, a^{test}\}} \prod_{j=1}^mP_{\theta}(a_{j}'|s_{j}',t_{i}').
\label{equ16}
\end{equation}
\vspace{-6 mm}
\section{Experiments and results}
\label{er}
\subsection{Datasets and experimental setup}
Two action datasets, MSRC-12 Kinect gesture~\cite{Fothergill2012} and 
WorkoutSU-10~\cite{Negin2013}, 
were used to evaluate our method. MSRC-12 Kinect gesture dataset is collected 
from 30 subjects performing 12 gestures and contains 594 video sequences of 
skeletal body part movements. Each sequence contains 9 to 11 instances. In 
total, there are 6244 instances in this dataset. WorkoutSu-10 dataset is 
collected from 15 subjects performing 10 fitness exercises.
Each exercise has been repeated 10 times by each 
subject. It contains 510550 frames resulting in 1500 instances and large time 
span for each instance. The large time spans make our method valuable in 
key frames extraction.

\begin{figure}[t]
\begin{center}

\begin{tabular}{cccccccc}
{\includegraphics[width = \figwidth mm,height = \figheight mm]{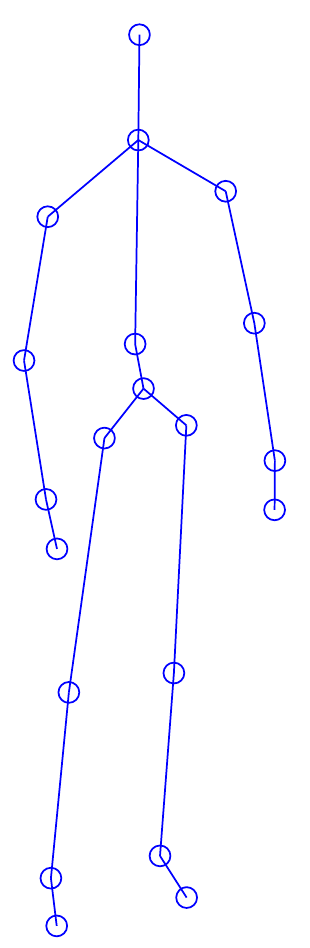}}&
{\includegraphics[width = \figwidth mm,height = \figheight mm]{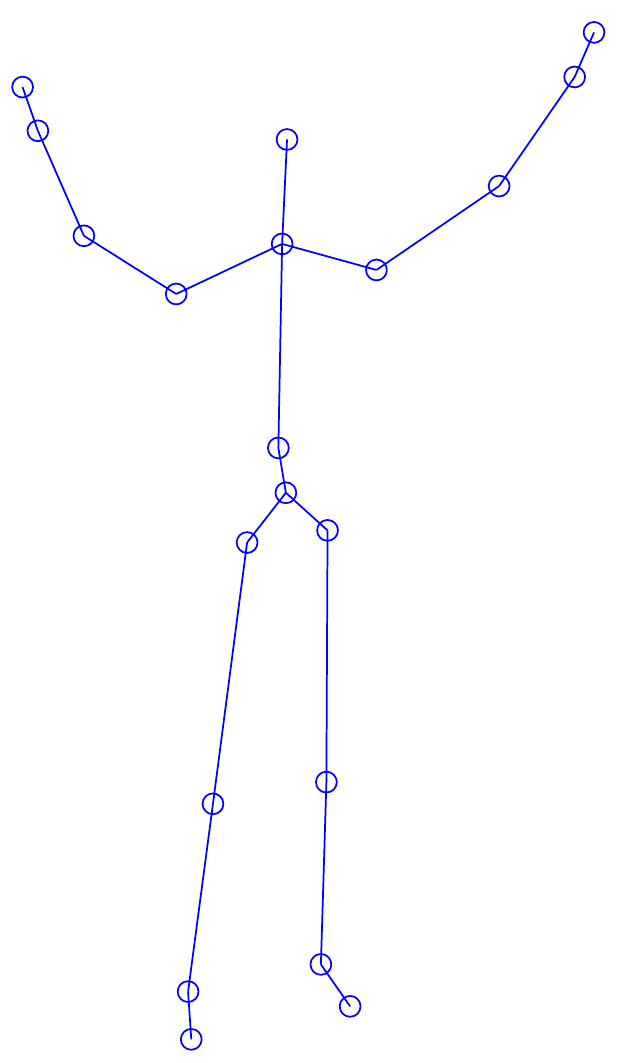}}&
{\includegraphics[width = \figwidth mm,height = \figheight mm]{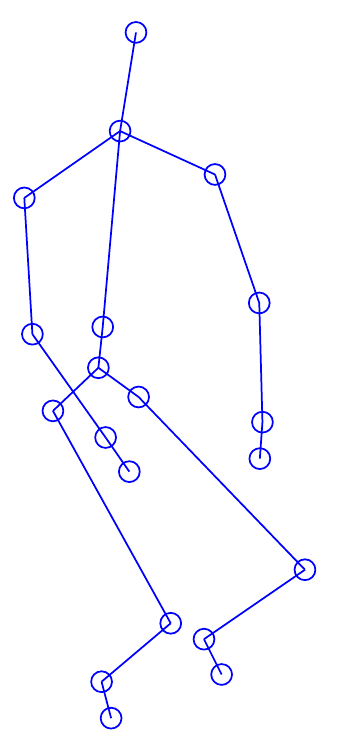}}&
{\includegraphics[width = \figwidth mm,height = \figheight mm]{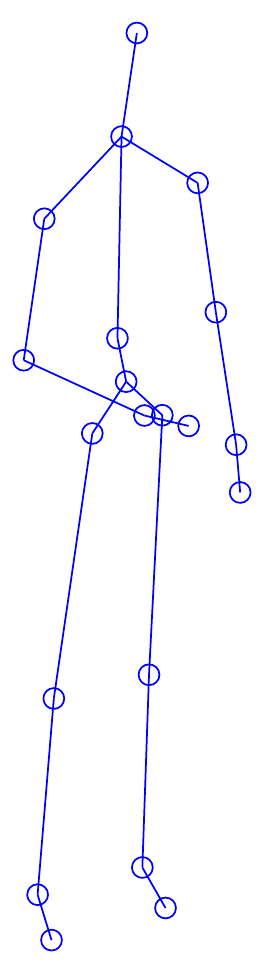}}&
{\includegraphics[width = \figwidth mm,height = \figheight mm]{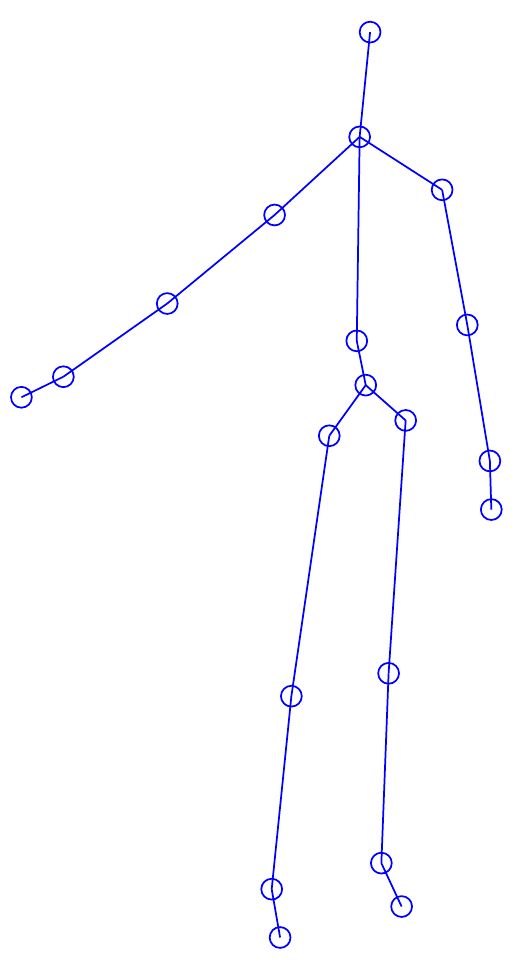}}&
{\includegraphics[width = \figwidth mm,height = \figheight mm]{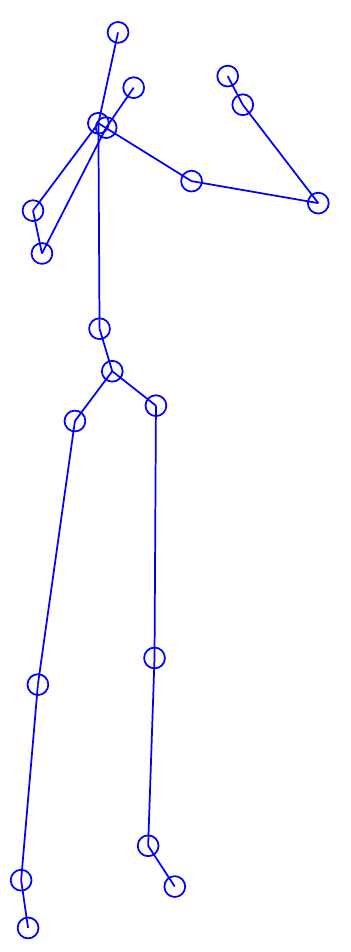}}&
{\includegraphics[width = \figwidth mm,height = \figheight mm]{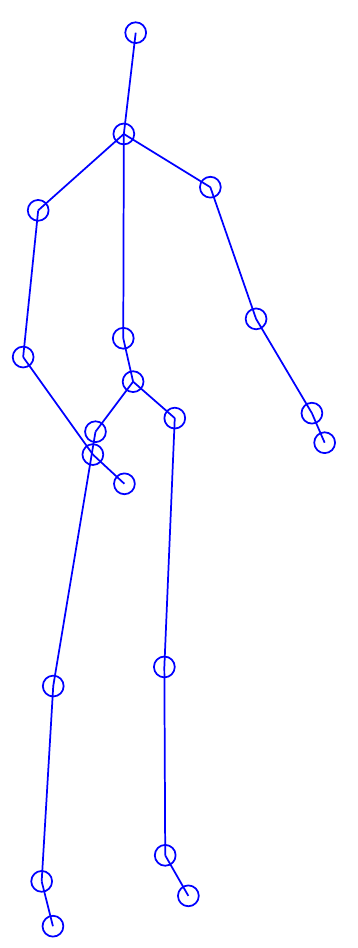}}&
{\includegraphics[width = \figwidth mm,height = \figheight mm]{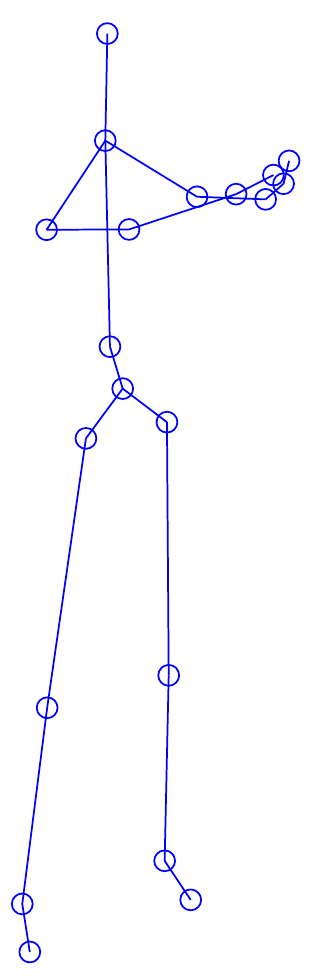}}\\
$T_1$&$T_{2}$&$T_3$&$T_{4}$&$T_5$&$T_{6}$&$T_7$&$T_{8}$\\
{\includegraphics[width = \figwidth mm,height = \figheight mm]{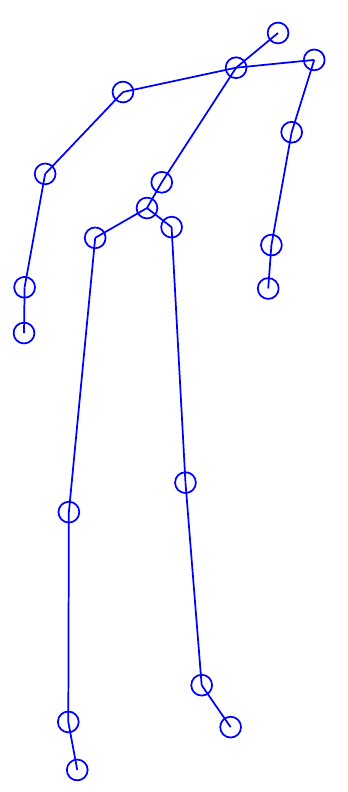}}&
{\includegraphics[width = \figwidth mm,height = \figheight mm]{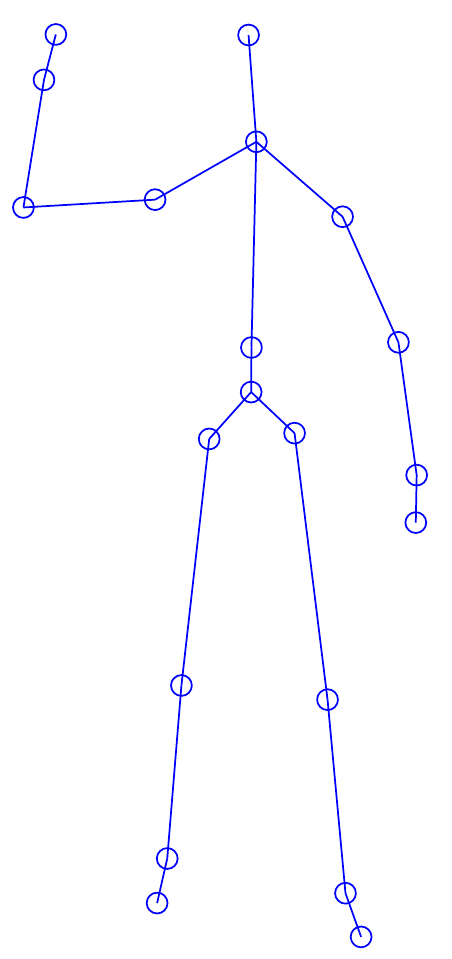}}&
{\includegraphics[width = \figwidth mm,height = \figheight mm]{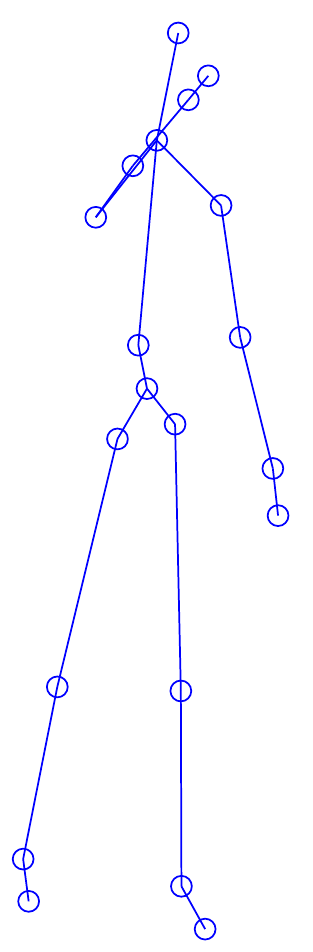}}&
{\includegraphics[width = \figwidth mm,height = \figheight mm]{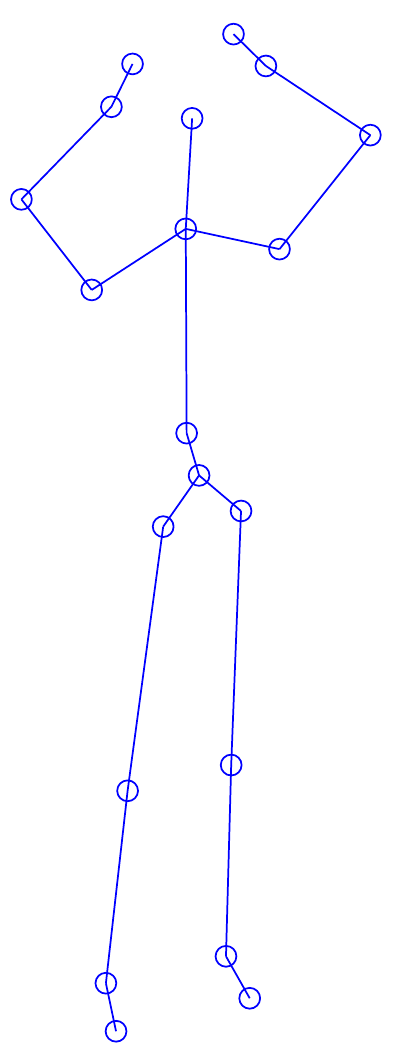}}&
{\includegraphics[width = \figwidth mm,height = \figheight mm]{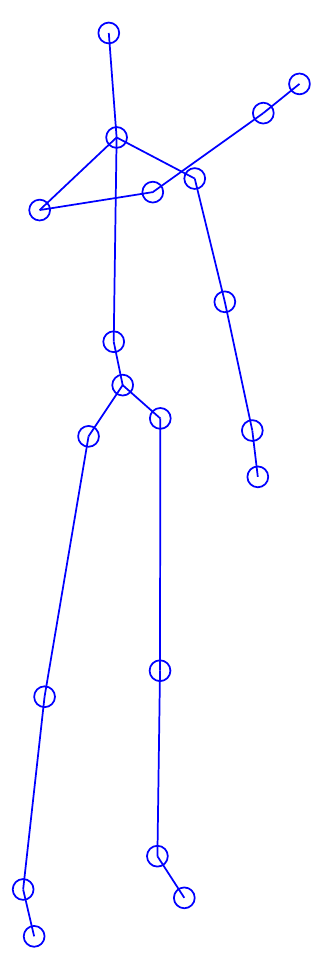}}&
{\includegraphics[width = \figwidth mm,height = \figheight mm]{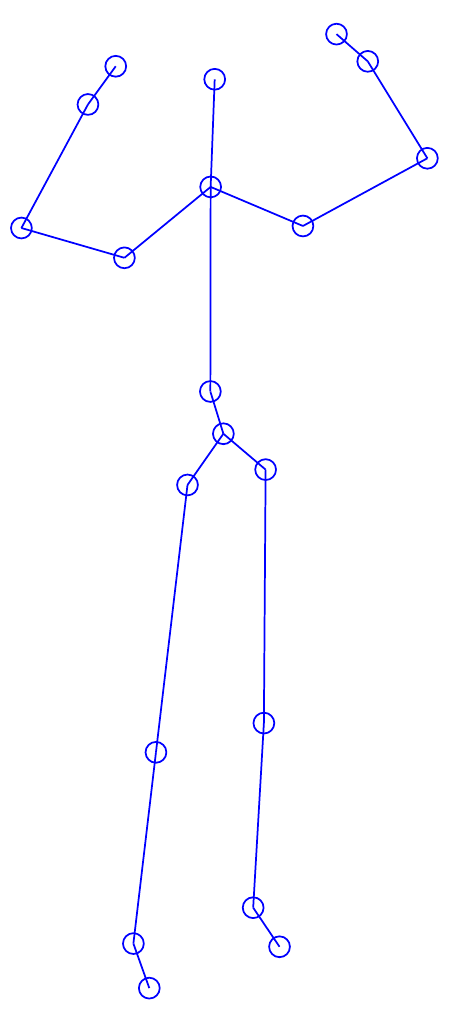}}&
{\includegraphics[width = \figwidth mm,height = \figheight mm]{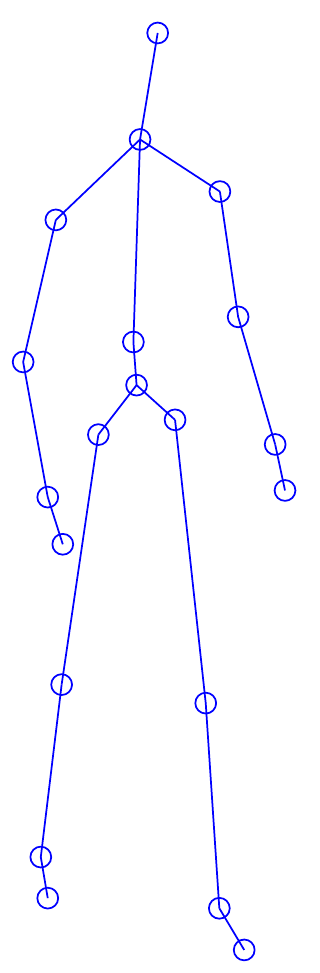}}&
{\includegraphics[width = \figwidth mm,height = \figheight mm]{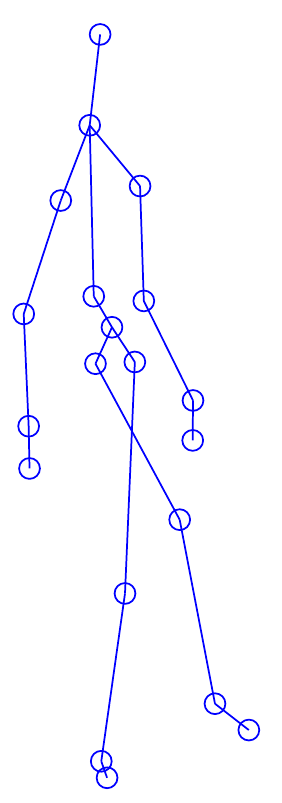}}\\
$T_9$&$T_{10}$&$T_{11}$&$T_{12}$&$T_{13}$&$T_{14}$&$T_{15}$&$T_{16}$\\
{\includegraphics[width = \figwidth mm,height = \figheight mm]{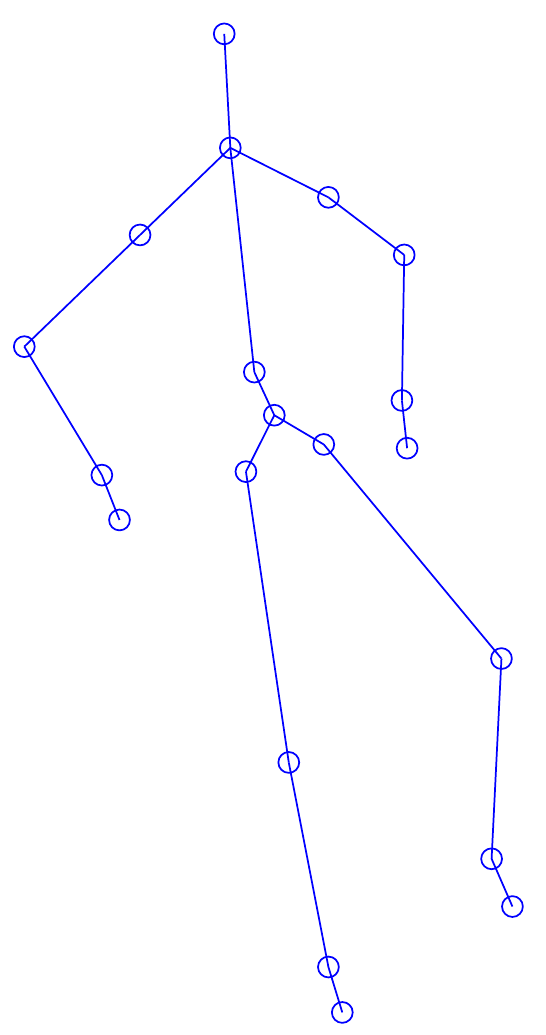}}&
{\includegraphics[width = \figwidth mm,height = \figheight mm]{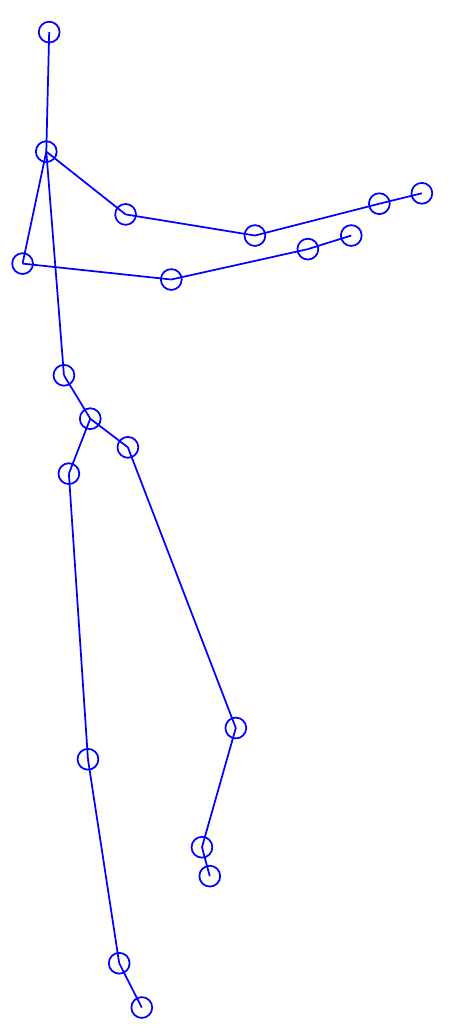}}&
{\includegraphics[width = \figwidth mm,height = \figheight mm]{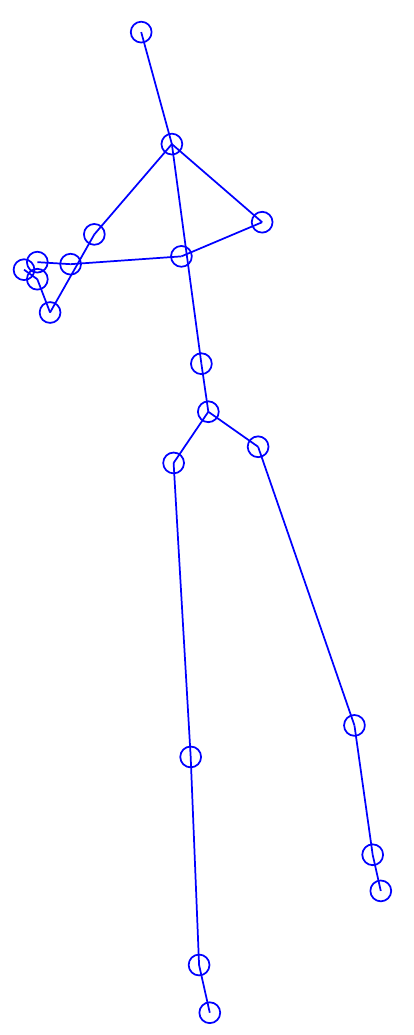}}&
{\includegraphics[width = \figwidth mm,height = \figheight mm]{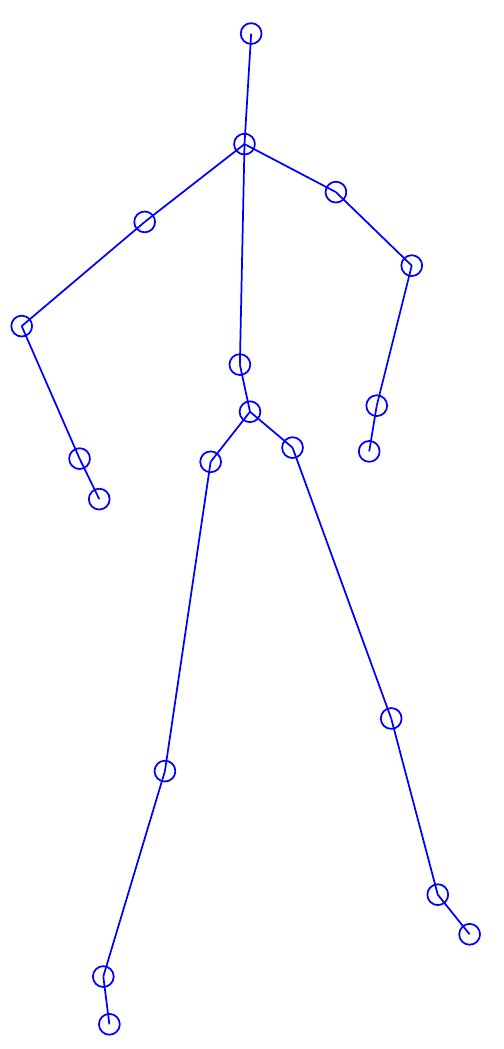}}&
{\includegraphics[width = \figwidth mm,height = \figheight mm]{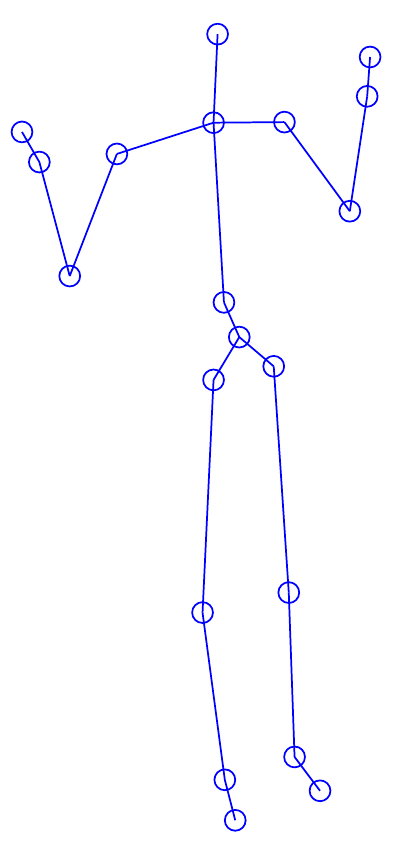}}&
{\includegraphics[width = \figwidth mm,height = \figheight mm]{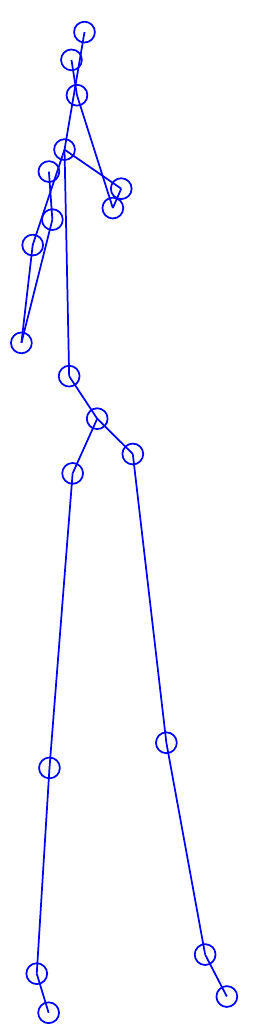}}&
{\includegraphics[width = \figwidth mm,height = \figheight mm]{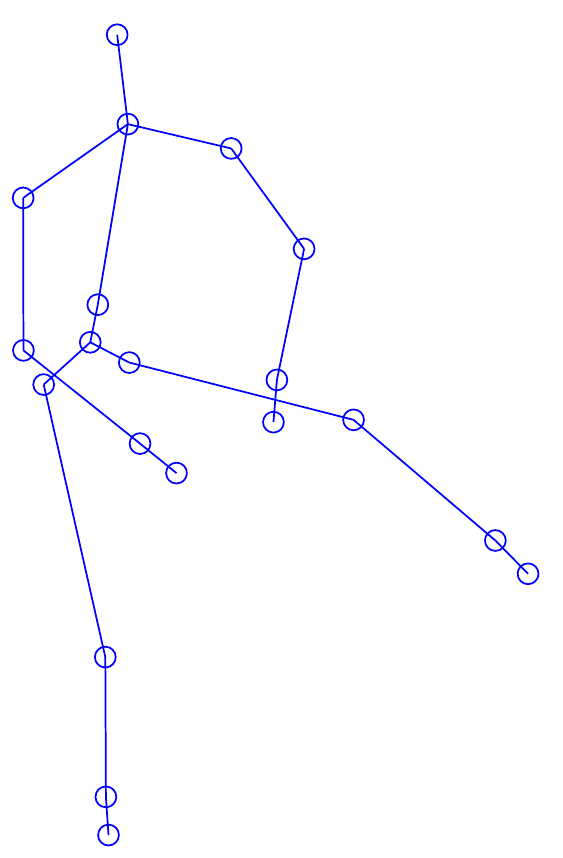}}&
{\includegraphics[width = \figwidth mm,height = \figheight mm]{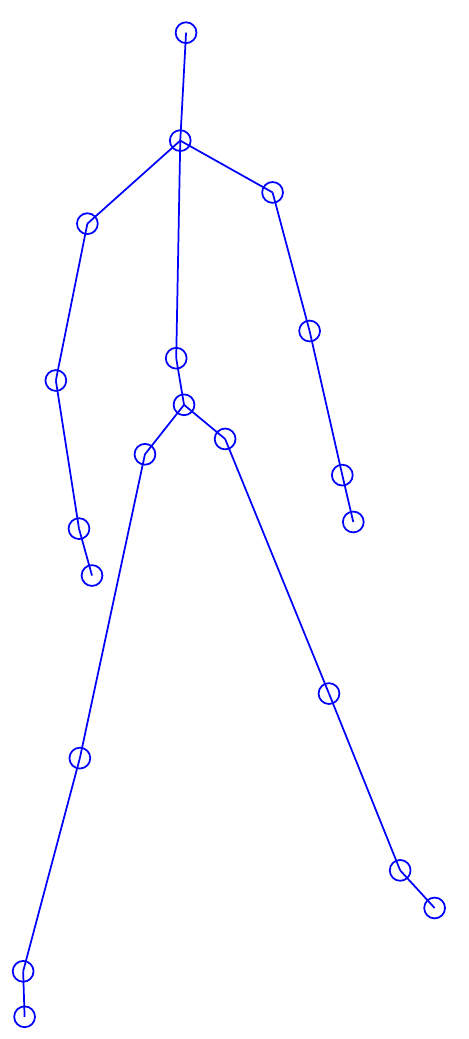}}\\

$T_{17}$&$T_{18}$&$T_{19}$&$T_{20}$&$T_{21}$&$T_{22}$&$T_{23}$&$T_{24}$\\

{\includegraphics[width = \figwidth mm,height = \figheight mm]{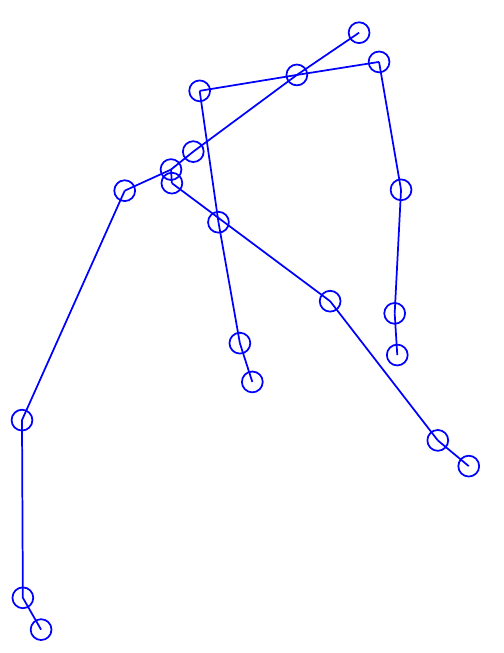}}&
{\includegraphics[width = \figwidth mm,height = \figheight mm]{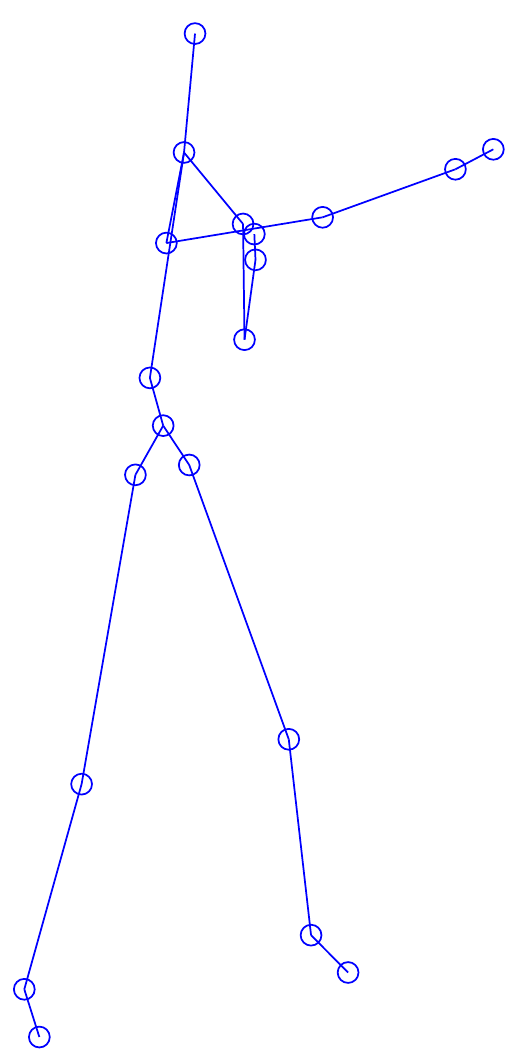}}&
{\includegraphics[width = \figwidth mm,height = \figheight mm]{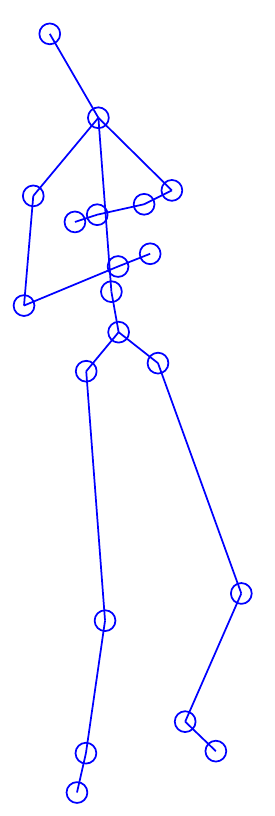}}&
{\includegraphics[width = \figwidth mm,height = \figheight mm]{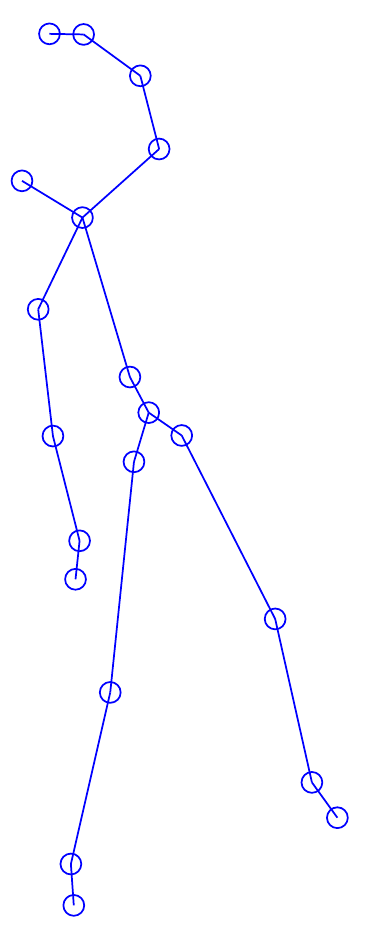}}&
&
&
&
\\
$T_{25}$&$T_{26}$&$T_{27}$&$T_{28}$&&&&\\
\end{tabular}
\end{center}
\vspace{-3 mm}
\caption{Ground truth optimal visual poses corresponding to semantic poses.}
\vspace{-3 mm}
\label{fig7_des} 
\end{figure}

Textual instructions of actions in the two datasets and linguistics 
description of extracted semantic poses are manually described which shown in supplementary material\footnote{\begin{footnotesize}
Supplementary material is attached at the end of paper.
\end{footnotesize}}. However, it is not difficult to automatically obtain textual instructions as they are often used by sports trainers and may be available online. In 
total, 16 and 15 semantic poses are respectively applied to MSRC-12 and 
WorkoutSu-10 datasets. The ground truth lexicon can be seen in 
Figure~\ref{fig7_des} which illustrates corresponding optimal visual pose of 
semantic pose. Here, symbols have been used to represent semantic poses.
​

Experiments were conducted to evaluate the proposed method on two
classification problems including trained and untrained actions. In order to 
compare with the state-of-the-art algorithms, cross-subject evaluation scheme 
was applied on the instance level of actions.  Moreover, the initial value of 
translation probability $P(S_p|T_q)$ is assigned a uniform probability based on 
$\sum_{S_p} P(S_p|T_q)=1$.
\vspace{-4 mm}
\subsection{Results}
\subsubsection{Trained action classification}
\begin{figure}[!t]
\begin{center}
\begin{tabular}{cccccccc}
{\includegraphics[width = \figwidth mm,height = \figheight mm]{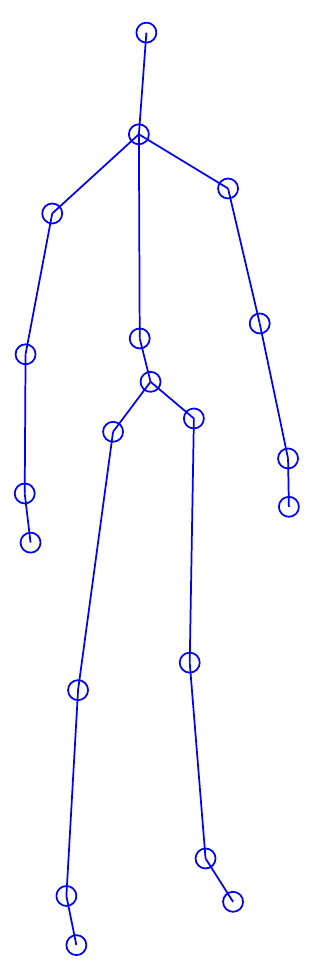}}&
{\includegraphics[width = \figwidth mm,height = \figheight mm]{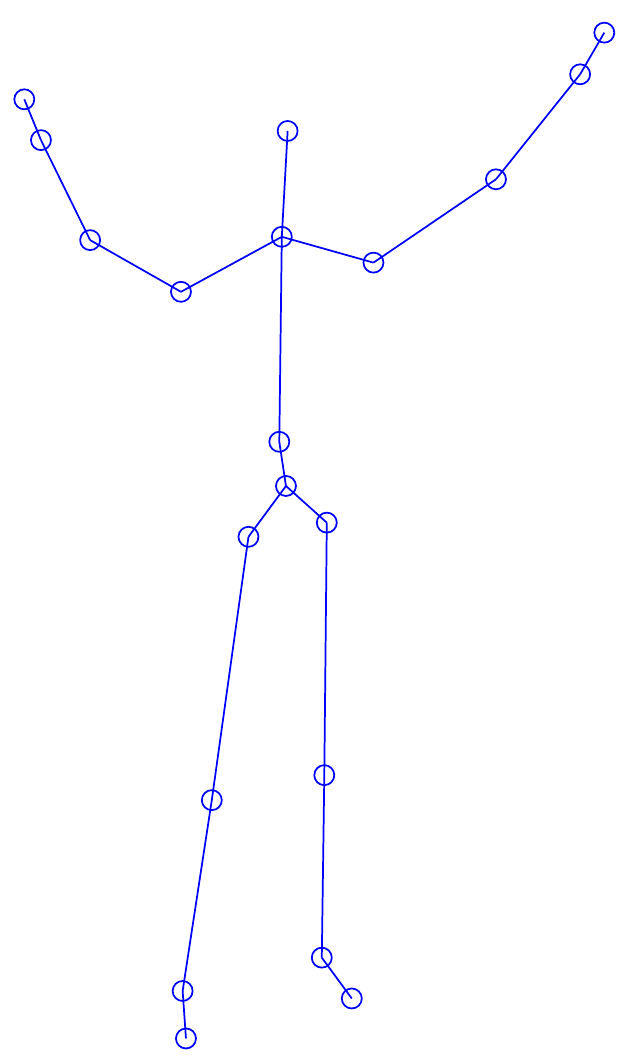}}&
{\includegraphics[width = \figwidth mm,height = \figheight mm]{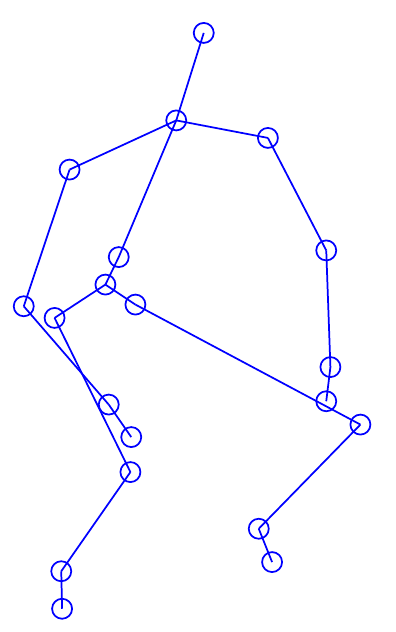}}&
{\includegraphics[width = \figwidth mm,height = \figheight mm]{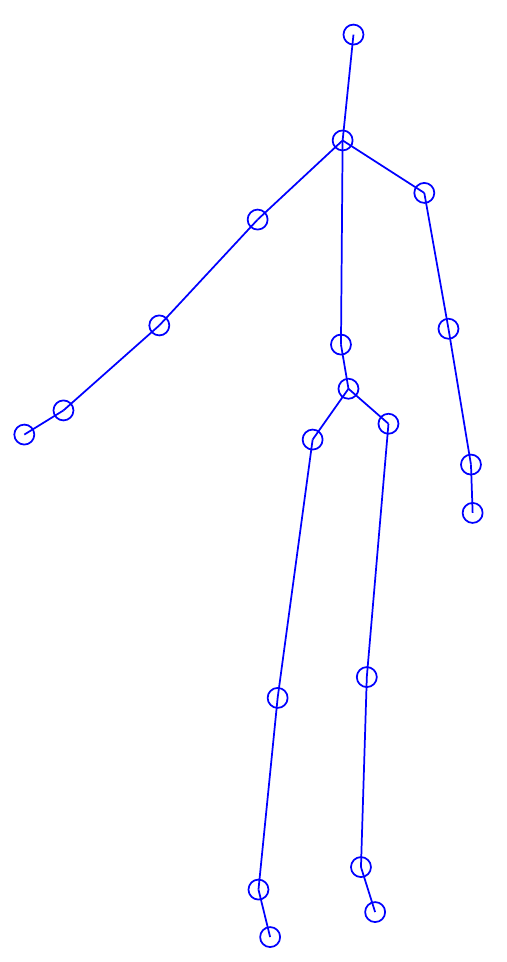}}&
{\includegraphics[width = \figwidth mm,height = \figheight mm]{./MSRC12keyposepoint/5-eps-converted-to.pdf}}&
{\includegraphics[width = \figwidth mm,height = \figheight mm]{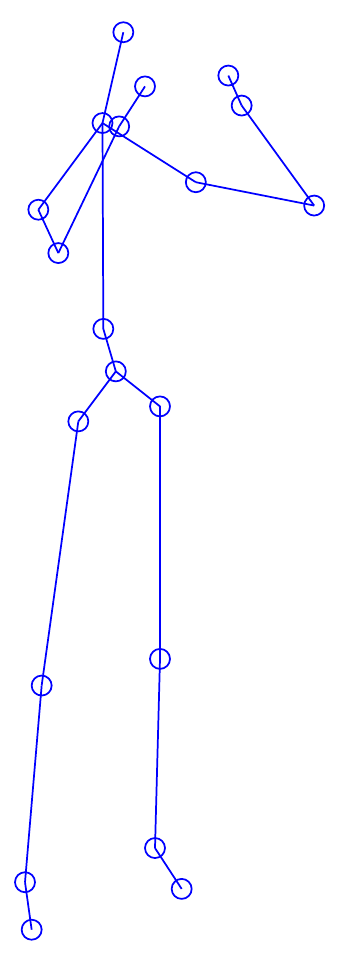}}&
{\includegraphics[width = \figwidth mm,height = \figheight mm]{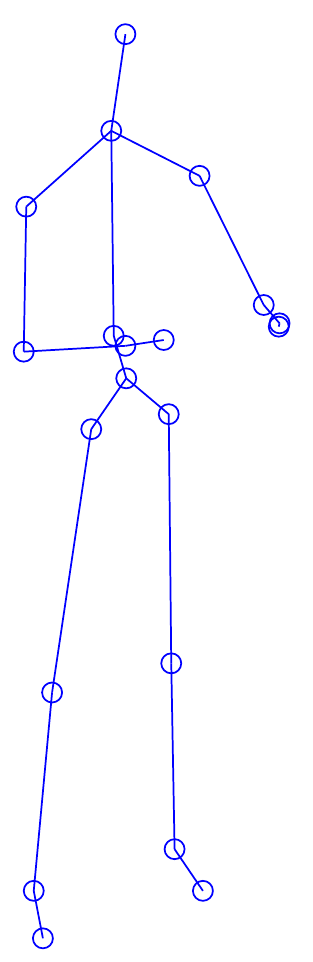}}&
{\includegraphics[width = \figwidth mm,height = \figheight mm]{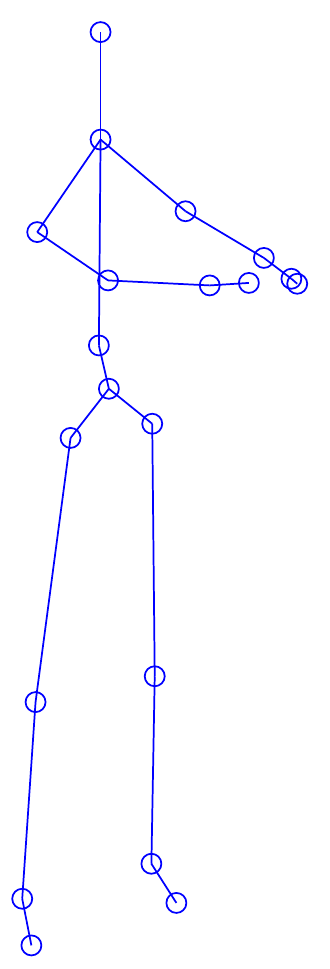}}\\
$T_1$&$T_{2}$&$T_3$&$T_{4}$&$T_5$&$T_{6}$&$T_7$&$T_{8}$\\
{\includegraphics[width = \figwidth mm,height = \figheight mm]{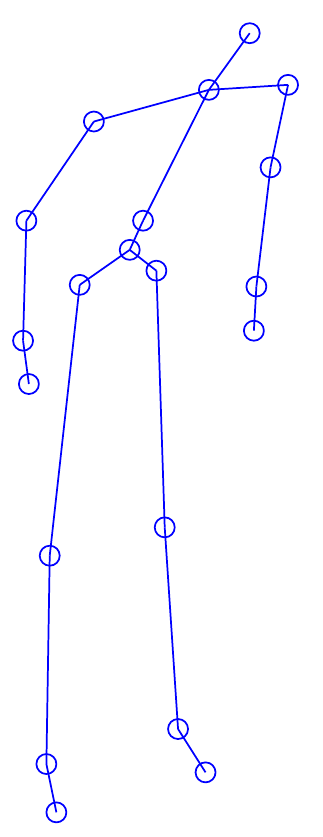}}&
{\includegraphics[width = \figwidth mm,height = \figheight mm]{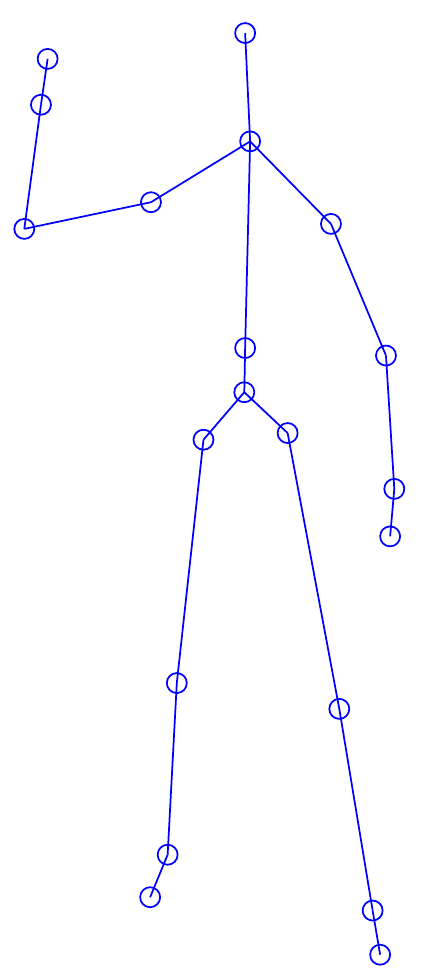}}&
{\includegraphics[width = \figwidth mm,height = \figheight mm]{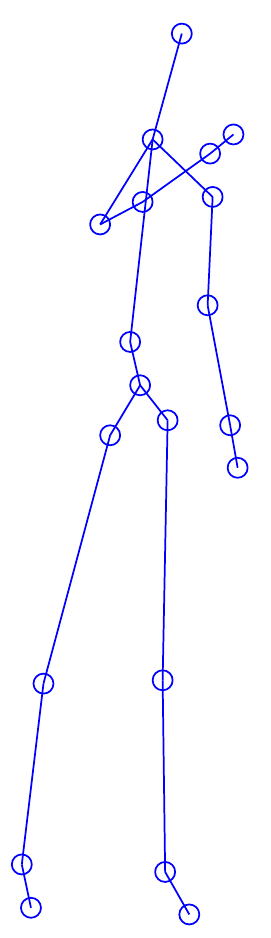}}&
{\includegraphics[width = \figwidth mm,height = \figheight mm]{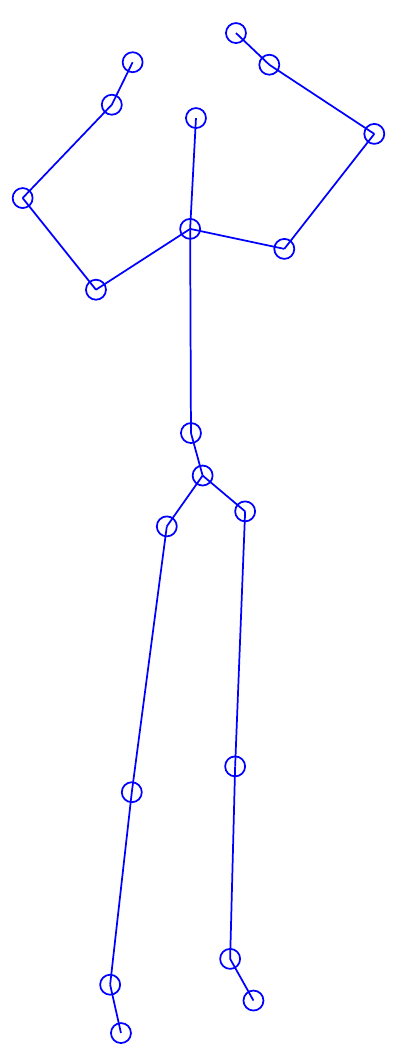}}&
{\includegraphics[width = \figwidth mm,height = \figheight mm]{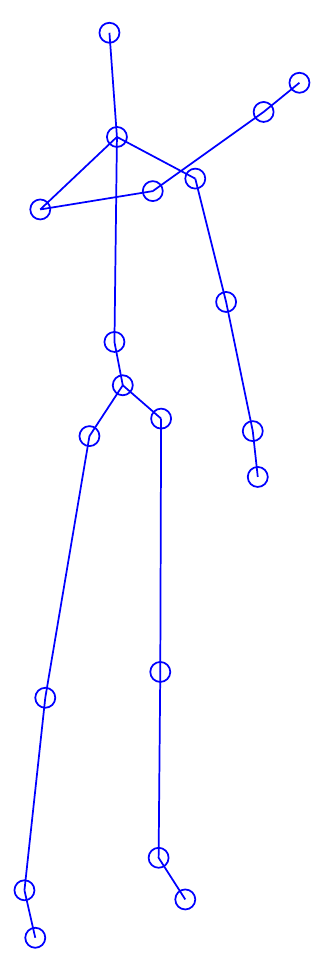}}&
{\includegraphics[width = \figwidth mm,height = \figheight mm]{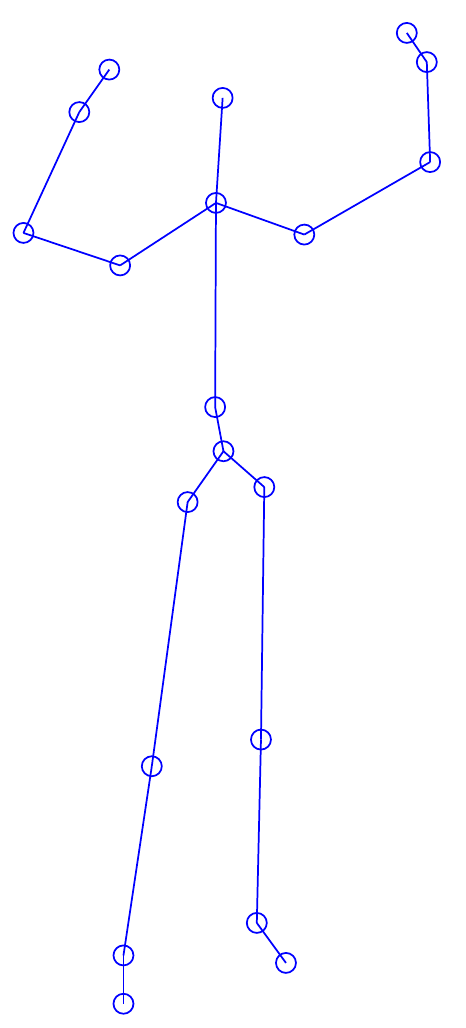}}&
{\includegraphics[width = \figwidth mm,height = \figheight mm]{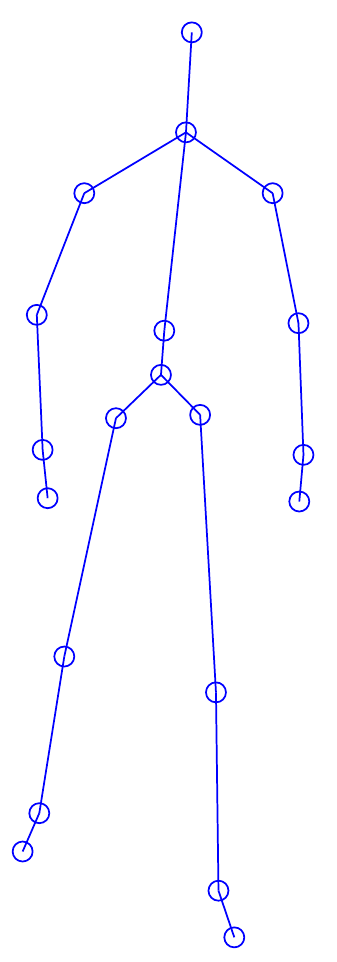}}&
{\includegraphics[width = \figwidth mm,height = \figheight mm]{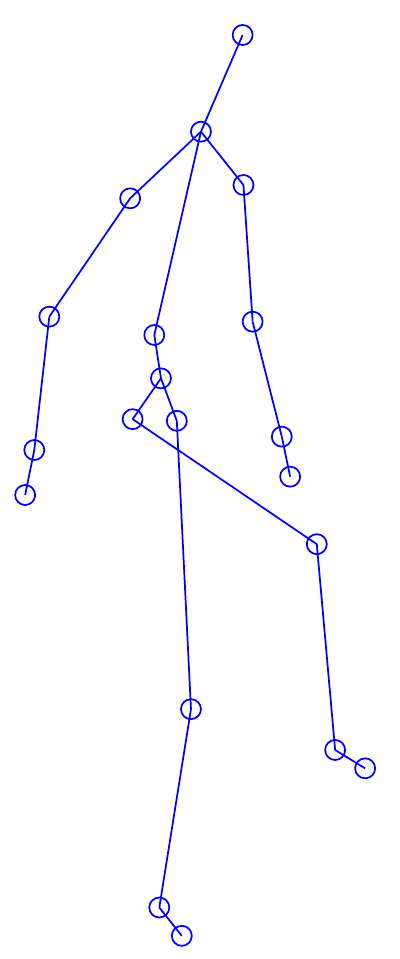}}\\
$T_9$&$T_{10}$&$T_{11}$&$T_{12}$&$T_{13}$&$T_{14}$&$T_{15}$&$T_{16}$\\
\end{tabular}
\end{center}
\vspace{-3 mm}
\caption{A lexicon based on MSRC-12 dataset.}
\label{fig3_msrc} 
\end{figure}

\begin{figure}[!t]
\begin{center}

\begin{tabular}{cccccccc}
{\includegraphics[width = \figwidth mm,height = \figheight mm]{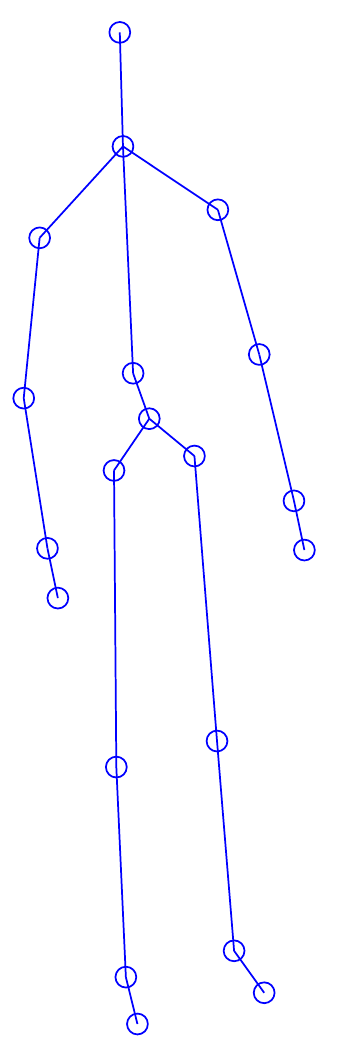}}&
{\includegraphics[width = \figwidth mm,height = \figheight mm]{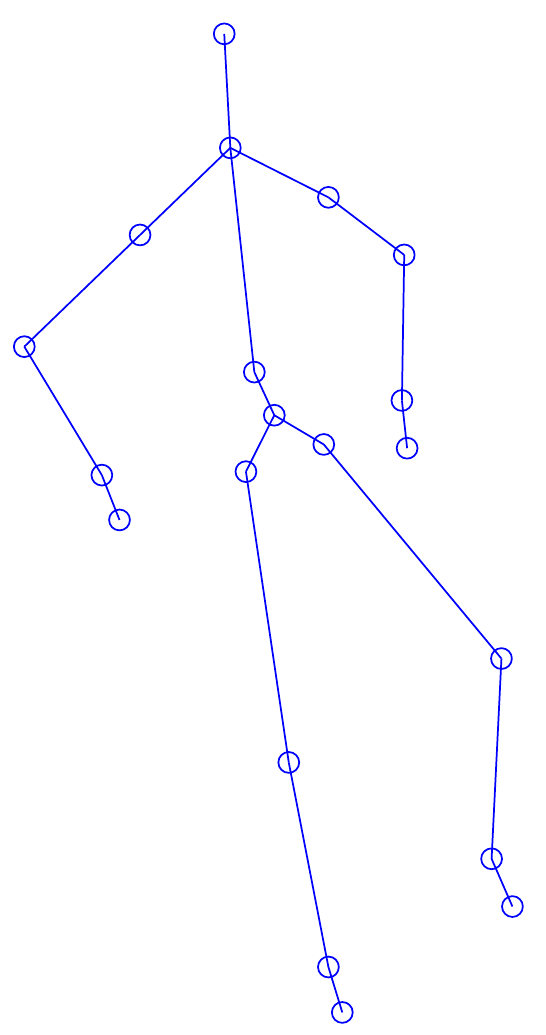}}&
{\includegraphics[width = \figwidth mm,height = \figheight mm]{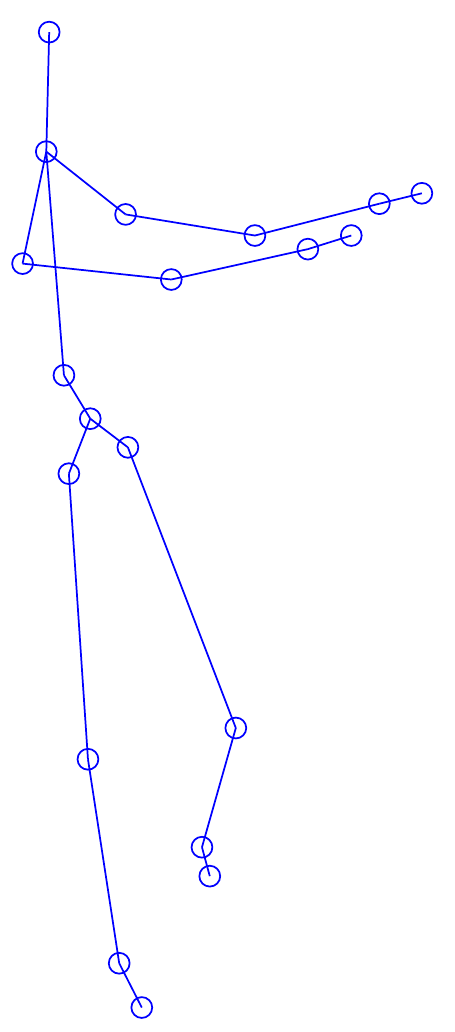}}&
{\includegraphics[width = \figwidth mm,height = \figheight mm]{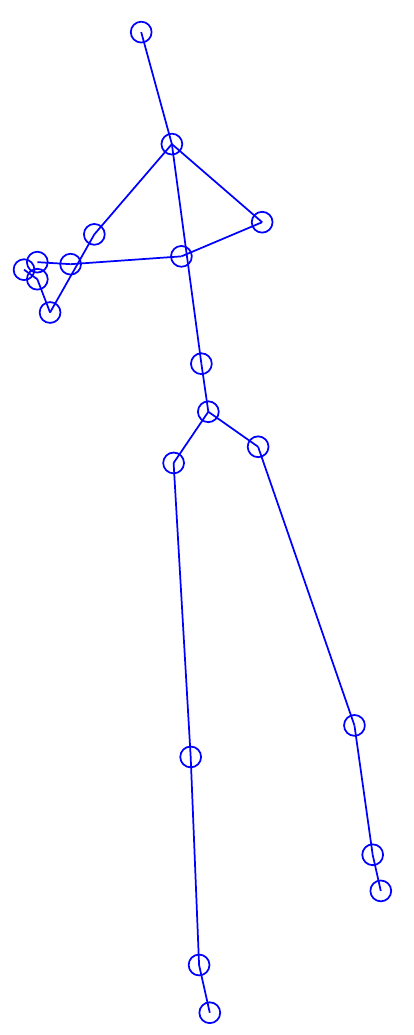}}&
{\includegraphics[width = \figwidth mm,height = \figheight mm]{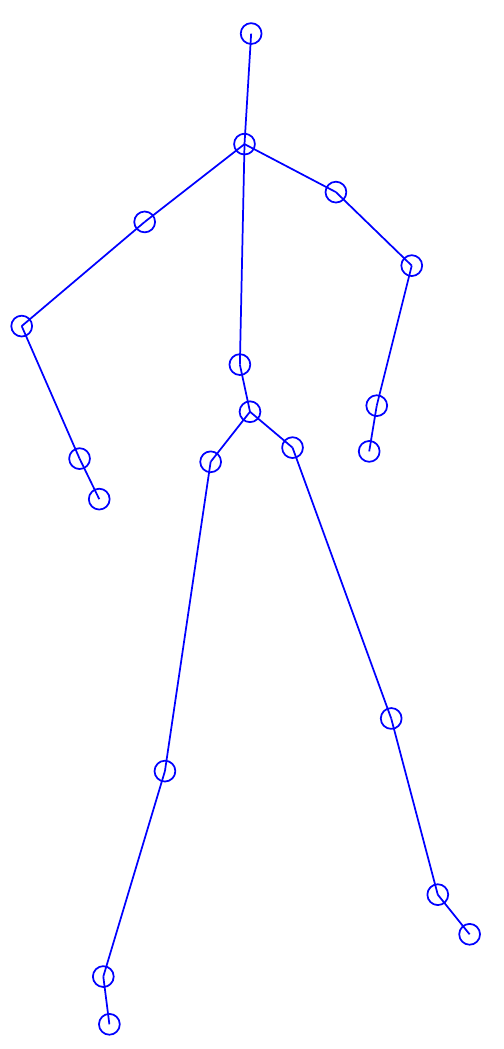}}&
{\includegraphics[width = \figwidth mm,height = \figheight mm]{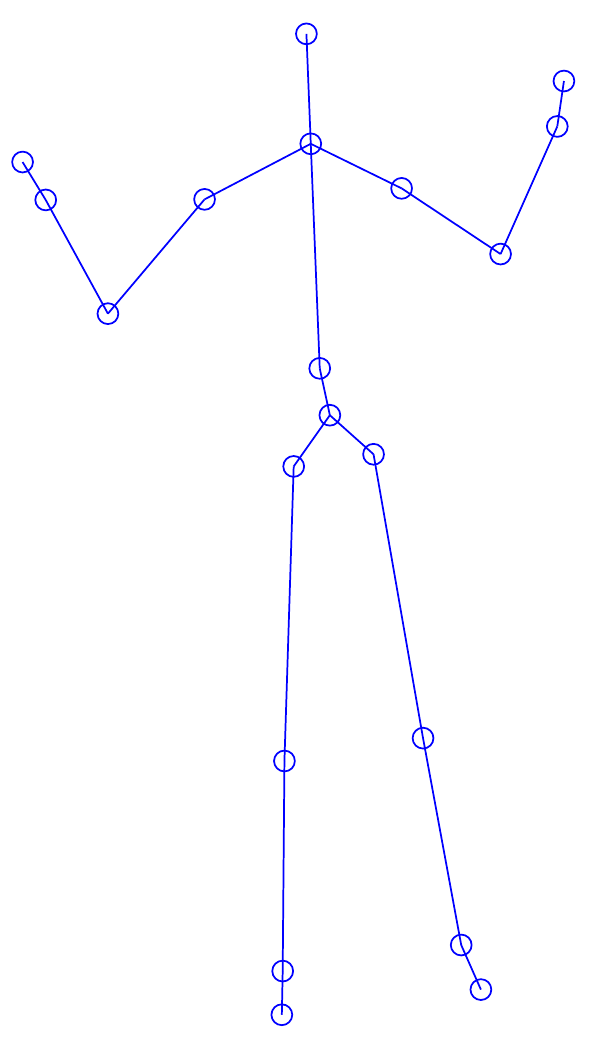}}&
{\includegraphics[width = \figwidth mm,height = \figheight mm]{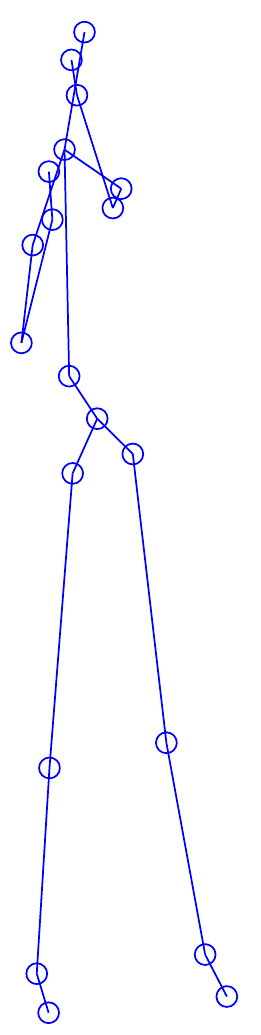}}&
{\includegraphics[width = \figwidth mm,height = \figheight mm]{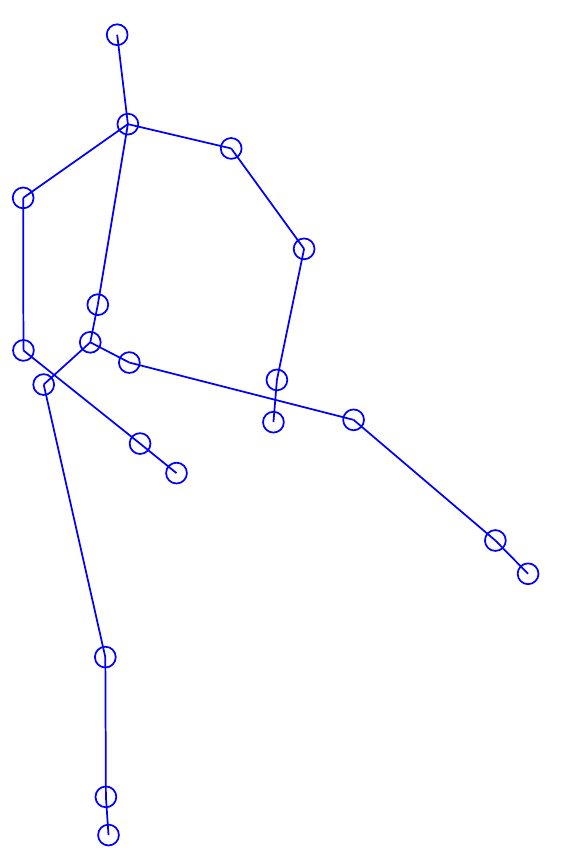}}\\
$T_{1}$&$T_{17}$&$T_{18}$&$T_{19}$&$T_{20}$&$T_{21}$&$T_{22}$&$T_{23}$\\

{\includegraphics[width = \figwidth mm,height = \figheight mm]{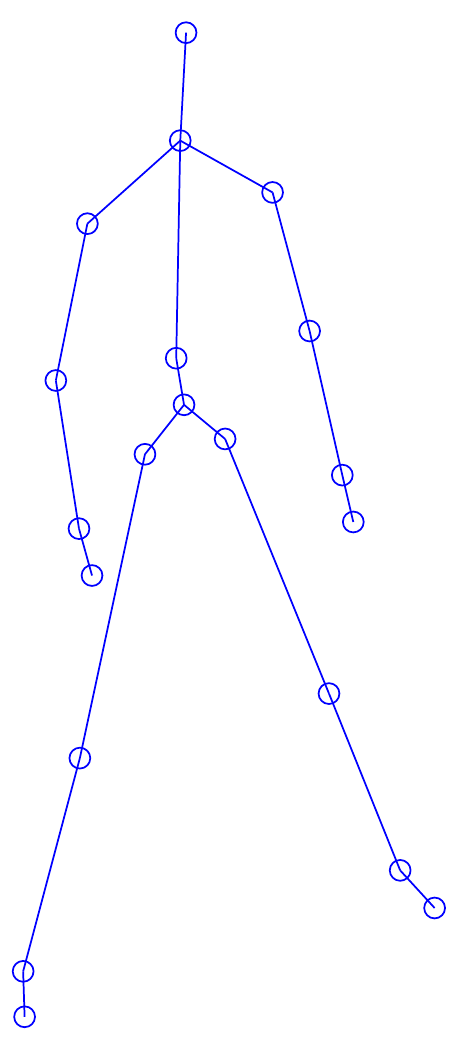}}&
{\includegraphics[width = \figwidth mm,height = \figheight mm]{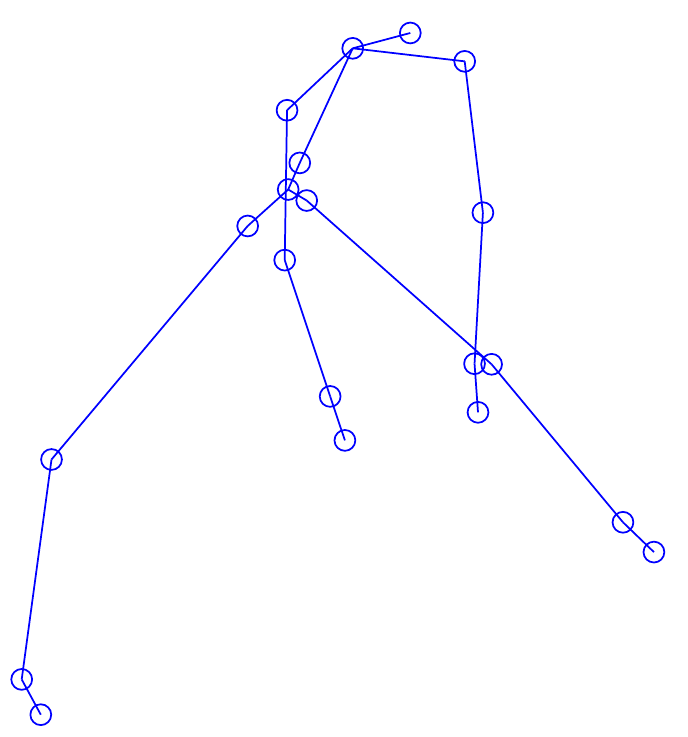}}&
{\includegraphics[width = \figwidth mm,height = \figheight mm]{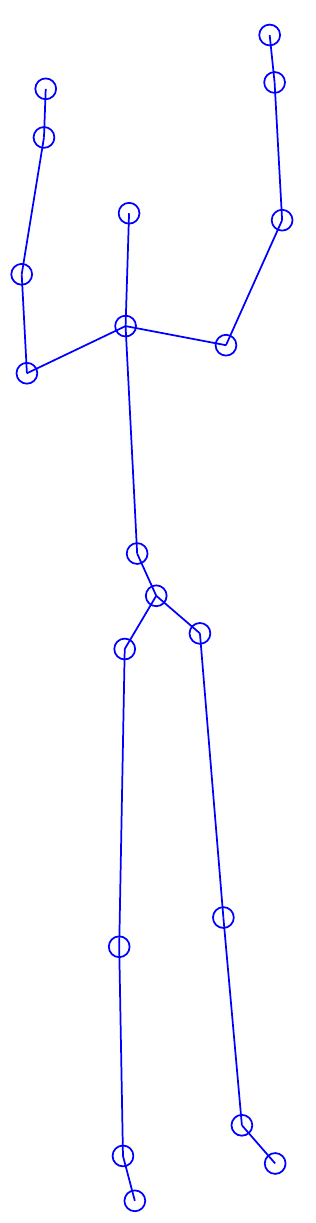}}&
{\includegraphics[width = \figwidth mm,height = \figheight mm]{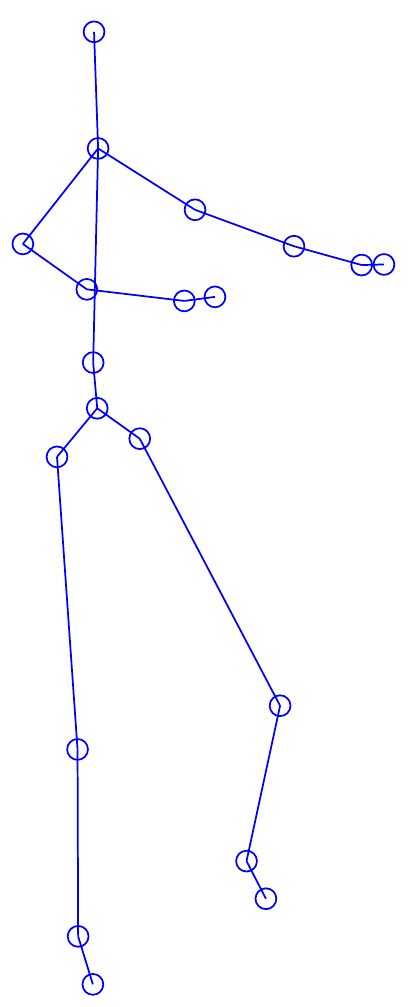}}&
{\includegraphics[width = \figwidth mm,height = \figheight mm]{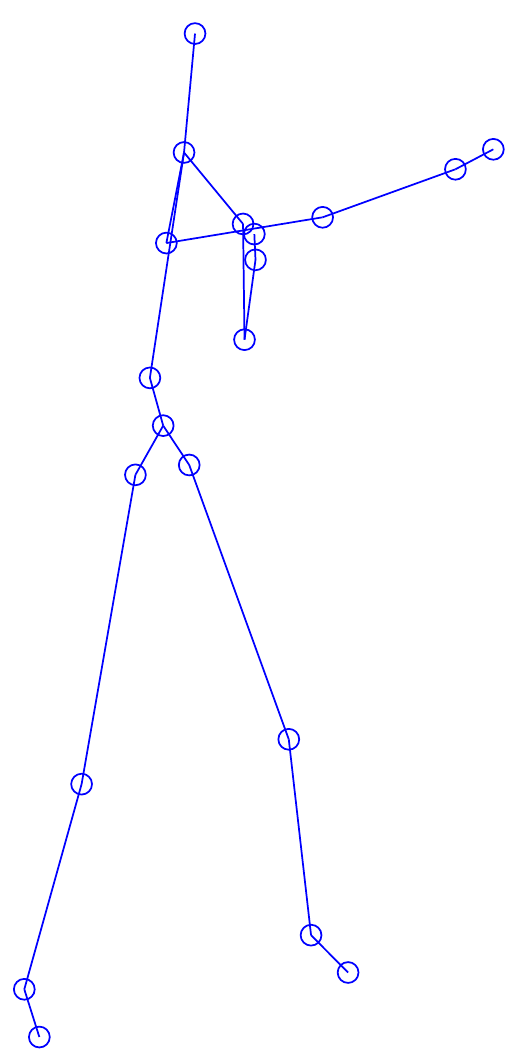}}&
{\includegraphics[width = \figwidth mm,height = \figheight mm]{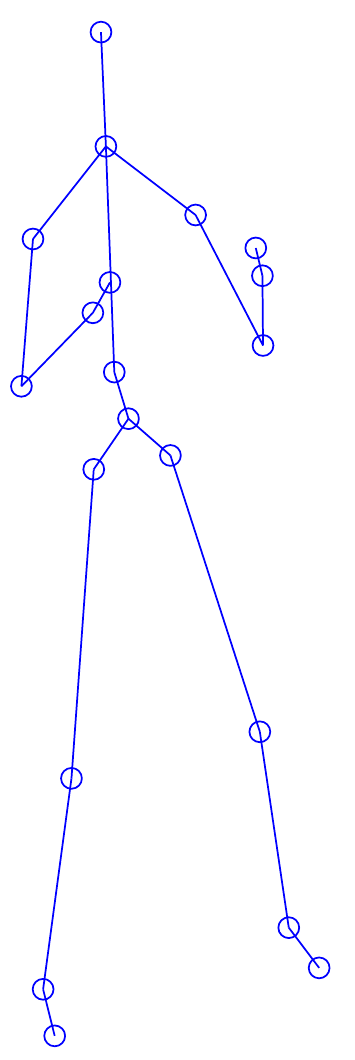}}&
{\includegraphics[width = \figwidth mm,height = \figheight mm]{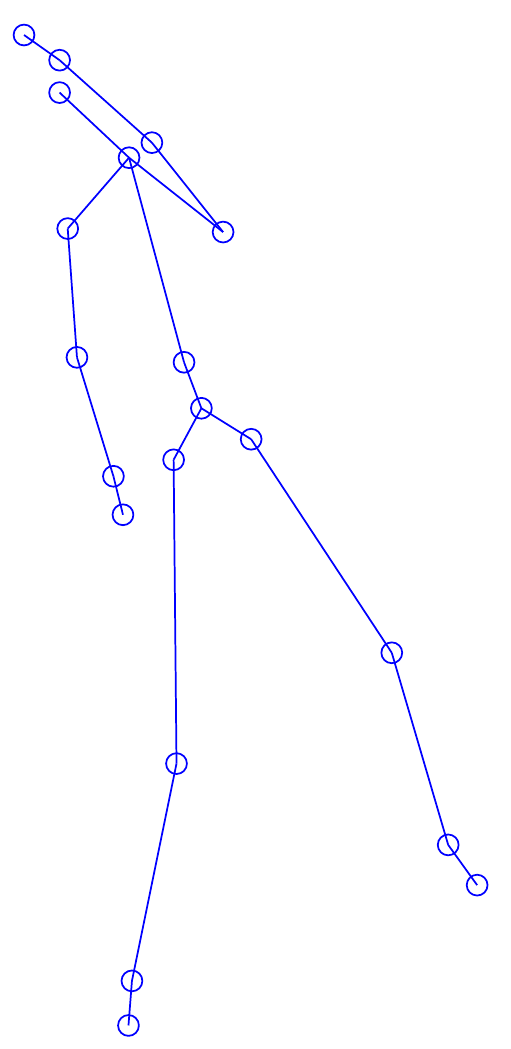}}&\\
$T_{24}$&$T_{25}$&$T_{2}$&$T_{3}$&$T_{26}$&$T_{27}$&$T_{28}$&\\

\end{tabular}
\end{center}
\vspace{-3 mm}
\caption{A lexicon based WorkoutSu-10 dataset.}
\label{fig5_workout} 
\end{figure}

We separately test two datasets and learned a lexicon (see 
Figure~\ref{fig3_msrc} and~\ref{fig5_workout}) for each
dataset used for action recognition. Comparing the learned lexicon 
with the ground truth lexicon, one finds the lexicon for MSRC-12 and 
WorkoutSu-10 are almost consistent with ground truth except the semantic pose 
$T_4$ in MSRC-12 dataset. Notice also that corresponding pose of 
semantic pose $T_4$ is confused with $T_5$. These two semantic poses are used 
by only one action ``Push right", which reduces the performance of the proposed 
method by fewer co-occurrence of semantic pose and visual pose candidates. The variation among 
subjects performing this action also results in confusion as subjects may ignore 
the elementary action from $T_1$ to $T_4$.

The learned lexicon was further verified through action recognition. The accuracy 
gained with MRSC-12 and WorkoutSu-10 dataset are respectively 85.86\% and 
98.71\%. Comparative results with discriminative models are shown in 
Table~\ref{tab_com}. Although the performance on MSRC-12 is slightly worse than 
discriminative models~\cite{Hussein2013,Negin2013,Zhou2014}, it is a reasonable 
result when considering the fact that we do not model reference object (G) and simply consider the whole body as object. Consequently, it is hard to distinguish actions ``Googles” and ``Had enough” which have particular reference objects. The result 
based on WorkoutSu-10 dataset outperforms the discriminative model 
RDF~\cite{Negin2013} even though the number of instances in~\cite{Negin2013} was 
300 fewer than in our experiment. 

To demonstrate the performance of semantic action recognition with the aid 
of textual instruction, we compare the proposed method with state-of-the-art 
semantic learning methods. As 
attributes~\cite{Liu2011,zhang2013attribute} are hard to describe and not 
comparable, we opt to compare it with state-of-the-art latent semantic 
learning. The comparisons are made with generative models and results are shown 
in Table~\ref{tab_com}; the proposed method can be catgorised as a 
generative model. In particular, we compare it with classical latent Dirichlet 
allocation (LDA) model~\cite{Niebles2008} and a hierarchical generative model 
(HGM)~\cite{yang2014hierarchical} which is a two-layer LDA. In both models, word is 
similar to visual pose candidate of the proposed method and topic is similar to semantic 
pose. Hence, in LDA the number of topics is equal to the number of semantic poses. In HGM, the number of global topics is same as the number of semantic 
poses. Comparative 
results show that the proposed method outperforms HGM and 
LDA, clearly indicating its semantic representation power for action 
recognition. Moreover, experiments investigating the role of $k$ (set size of 
visual pose candidate) showed an increased rate for the proposed method 
with increased value of $k$, levelling off when $k$ is about 5 or 6 
times as many as the size of semantic pose set \footnote{\begin{footnotesize}
Graphical result is available 
in the supplementary material.
\end{footnotesize}}. 

​

​

\begin{table}[!t]
\centering
\caption{ Comparative results: trained action classification.}
\begin{tabular}{|l|c|c|}
\hline 
Methods &MSRC-12& WorkoutSu-10 \\
\hline
Cov3DJ~\cite{Hussein2013}& 91.70\%&-  \\
RDF~\cite{Negin2013}&\textbf{94.03\%}& 98\%\\
ELC-KSVD~\cite{Zhou2014}&90.22\%&-\\
LDA\cite{Niebles2008}&74.81\%&92.27\%\\
HGM~\cite{yang2014hierarchical}& 66.25\%&82.37\%\\
Ours &85.86\%\ &\textbf{98.71\%}\\
 \hline
\end{tabular}
\label{tab_com}
\end{table}
\vspace{-10 mm}
\subsubsection{Zero-shot classification}
Zero-shot classification is to recognize an action that has not been trained before. Notice that the four semantic poses are shared in different 
actions among two datasets including $T_1$, $T_2$, $T_3$ and $T_7$. This 
motivated the selection of all actions that used these semantic poses for 
experiments. We also constructed a synthesized set of composite activities by 
concatenating single actions from WorkoutSu-10 in order to enlarge the number of 
test actions. Specifically, training actions include: \textit{Cheap weapon, 
Beat both, Hip flexion ($A_1$), Trunk rotation ($A_2$), Hip adductor 
stretch ($B_2$), Hip adductor stretch ($B_3$), Curl-to-press ($C_1$) and 
Squats ($C_2$)}. Three 
single actions and three composite activities are used for testing and 
recognition accuracy is shown in Table~\ref{tab5}. Results demonstrate that the 
proposed method based on action semantic poses is effective in zero-shot action 
recognition.

\begin{table}[!t]
\centering
\caption{ Results of zero-shot classification.}
\begin{tabular}{|l|c|c|}
\hline 
\multicolumn{2}{|c|}{Testing set}&Accuracy\\
\hline
\multirow{ 3}{*}{Single action}&Lift arms& 92.72\%  \\
&Duck  &61.40\%\\
&Wind up &54.70\%\ \\
\hline
\multirow{ 3}{*}{Composite activity}&$A_1$ then $A_2$ &99.30\%\\
&$B_2$ then $B_3$ &100\%\\
&$C_1$ then $C_2$&98.67\%\\
\hline
\end{tabular}
\label{tab5}
\end{table}

 \vspace{-0.2cm}
\section{Conclusion}
 \vspace{-0.2cm}
\label{c}
This paper has presented a novel method of learning a pose lexicon that consists 
of semantic poses and visual poses. Experimental results showed that the
proposed method can effectively learn the mapping between semantic and visual 
poses and was verified in both pre-trained and zero-shot action 
recognition. 

The proposed method can be easily extended to 
semantic action recognition based on RGB or depth datasets and provides a 
foundation to build action-verb and activity-phrase hierarchies. A unique large 
lexicon can also be learned for action recognition involving different 
datasets. In addition, the future work about representing semantic poses will be explored to improve the proposed method through modelling reference objects and middle part of trajectories since we only considered start and end positions of trajectories.

{\footnotesize
\bibliographystyle{IEEEtran}
\bibliography{subref}

\begin{thebibliography}{10}
\providecommand{\url}[1]{#1}
\csname url@samestyle\endcsname
\providecommand{\newblock}{\relax}
\providecommand{\bibinfo}[2]{#2}
\providecommand{\BIBentrySTDinterwordspacing}{\spaceskip=0pt\relax}
\providecommand{\BIBentryALTinterwordstretchfactor}{4}
\providecommand{\BIBentryALTinterwordspacing}{\spaceskip=\fontdimen2\font plus
\BIBentryALTinterwordstretchfactor\fontdimen3\font minus
  \fontdimen4\font\relax}
\providecommand{\BIBforeignlanguage}[2]{{%
\expandafter\ifx\csname l@#1\endcsname\relax
\typeout{** WARNING: IEEEtran.bst: No hyphenation pattern has been}%
\typeout{** loaded for the language `#1'. Using the pattern for}%
\typeout{** the default language instead.}%
\else
\language=\csname l@#1\endcsname
\fi
#2}}
\providecommand{\BIBdecl}{\relax}
\BIBdecl

\bibitem{li2008expandable}
W.~Li, Z.~Zhang, and Z.~Liu, ``Expandable data-driven graphical modeling of
  human actions based on salient postures,'' \emph{IEEE Trans. Circuits Syst.
  Video Technol.}, vol.~18, no.~11, pp. 1499--1510, 2008.

\bibitem{weinland2011survey}
D.~Weinland, R.~Ronfard, and E.~Boyer, ``A survey of vision-based methods for
  action representation, segmentation and recognition,'' \emph{Comput. Vision
  Image Underst.}, vol. 115, no.~2, pp. 224--241, 2011.

\bibitem{wang2014mining}
P.~Wang, W.~Li, P.~Ogunbona, Z.~Gao, and H.~Zhang, ``Mining mid-level features
  for action recognition based on effective skeleton representation,'' in
  \emph{Proc. Int. Conf. Digital lmage Computing Techniques Appl.}, 2014.

\bibitem{wang2015convnets}
P.~Wang, W.~Li, Z.~Gao, C.~Tang, J.~Zhang, and P.~Ogunbona, ``Convnets-based
  action recognition from depth maps through virtual cameras and
  pseudocoloring,'' in \emph{Proc. Annual Conf. ACM Multimedia}, 2015.

\bibitem{wang2015action}
P.~Wang, W.~Li, Z.~Gao, J.~Zhang, C.~Tang, and P.~O. Ogunbona, ``Action
  recognition from depth maps using deep convolutional neural networks,''
  \emph{IEEE Trans. Human-Mach. Syst.}, Accepted 2 November 2015.

\bibitem{li2010action}
W.~Li, Z.~Zhang, and Z.~Liu, ``Action recognition based on a bag of 3{D}
  points,'' in \emph{Proc. Comput. Vision Pattern Recognit. Workshops}, 2010.

\bibitem{wang2013approach}
C.~Wang, Y.~Wang, and A.~L. Yuille, ``An approach to pose-based action
  recognition,'' in \emph{Proc. Comput. Vision Pattern Recognit.}, 2013.

\bibitem{eweiwi2014efficient}
A.~Eweiwi, M.~S. Cheema, C.~Bauckhage, and J.~Gall, ``Efficient pose-based
  action recognition,'' in \emph{Proc. Asian Conf. Comput. Vision}.\hskip 1em
  plus 0.5em minus 0.4em\relax Springer, 2014, pp. 428--443.

\bibitem{Talmy2003}
L.~Talmy, \emph{Toward a cognitive semantics}, M.~I. of~Technology, Ed.\hskip
  1em plus 0.5em minus 0.4em\relax MIT press, 2003, vol.~1.

\bibitem{petrov2006learning}
S.~Petrov, L.~Barrett, R.~Thibaux, and D.~Klein, ``Learning accurate, compact,
  and interpretable tree annotation,'' in \emph{Proc. Annul Meeting Assoc.
  Comput. Ling.}, 2006.

\bibitem{Fothergill2012}
S.~Fothergill, H.~Mentis, P.~Kohli, and S.~Nowozin, ``Instructing people for
  training gestural interactive systems,'' in \emph{Proc. SIGCHI Conf. Human
  Factor Comput. Syst.}, 2012.

\bibitem{Negin2013}
F.~Negin, F.~{\"O}zdemir, C.~B. Akg{\"u}l, K.~A. Y{\"u}ksel, and
  A.~Er{\c{c}}il, ``A decision forest based feature selection framework for
  action recognition from {RGB}-{D}epth cameras,'' in \emph{Image Anal. and
  Recognit.}\hskip 1em plus 0.5em minus 0.4em\relax Springer, 2013, pp.
  648--657.

\bibitem{laptev2008learning}
I.~Laptev, M.~Marsza{\l}ek, C.~Schmid, and B.~Rozenfeld, ``Learning realistic
  human actions from movies,'' in \emph{Proc. Comput. Vision Pattern
  Recognit.}, 2008.

\bibitem{Niebles2008}
J.~C. Niebles, H.~Wang, and L.~Fei-Fei, ``Unsupervised learning of human action
  categories using spatial-temporal words,'' \emph{Int. J. Comput. Vision},
  vol.~79, no.~3, pp. 299--318, 2008.

\bibitem{Wang2009}
Y.~Wang and G.~Mori, ``Human action recognition by semi-latent topic models,''
  \emph{IEEE Trans. Pattern Anal. Mach. Intell.}, vol.~31, no.~10, pp.
  1762--1774, 2009.

\bibitem{Hofmann2001}
T.~Hofmann, ``Unsupervised learning by probabilistic latent semantic
  analysis,'' \emph{Mach. Learning}, vol.~42, no. 1-2, pp. 177--196, 2001.

\bibitem{Blei2003}
D.~M. Blei, A.~Y. Ng, and M.~I. Jordan, ``Latent dirichlet allocation,''
  \emph{J. Mach. Learn Res.}, vol.~3, pp. 993--1022, 2003.

\bibitem{yang2014hierarchical}
S.~Yang, C.~Yuan, W.~Hu, and X.~Ding, ``A hierarchical model based on latent
  dirichlet allocation for action recognition,'' in \emph{Proc. Int. Conf.
  Pattern Recognit.}, 2014.

\bibitem{Liu2011}
J.~Liu, B.~Kuipers, and S.~Savarese, ``Recognizing human actions by
  attributes,'' in \emph{Proc. Comput. Vision Pattern Recognit.}, 2011.

\bibitem{zhang2013attribute}
Z.~Zhang, C.~Wang, B.~Xiao, W.~Zhou, and S.~Liu, ``Attribute regularization
  based human action recognition,'' \emph{IEEE Trans. Inf. Forensics Security},
  vol.~8, no.~10, pp. 1600--1609, 2013.

\bibitem{rohrbach2012script}
M.~Rohrbach, M.~Regneri, M.~Andriluka, S.~Amin, M.~Pinkal, and B.~Schiele,
  ``Script data for attribute-based recognition of composite activities,'' in
  \emph{Proc. European Conf. Comput. Vision}, 2012.

\bibitem{guadarrama2013youtube2text}
S.~Guadarrama, N.~Krishnamoorthy, G.~Malkarnenkar, S.~Venugopalan, R.~Mooney,
  T.~Darrell, and K.~Saenko, ``Youtube2text: Recognizing and describing
  arbitrary activities using semantic hierarchies and zero-shot recognition,''
  in \emph{Proc. Int. Conf. Comput. Vision}, 2013.

\bibitem{Duygulu2002}
P.~Duygulu, K.~Barnard, J.~F. de~Freitas, and D.~A. Forsyth, ``Object
  recognition as machine translation: {L}earning a lexicon for a fixed image
  vocabulary,'' in \emph{Proc. European Conf. Comput. Vision}, 2002.

\bibitem{Zanfir2013}
M.~Zanfir, M.~Leordeanu, and C.~Sminchisescu, ``The moving pose: {An} efficient
  3{D} kinematics descriptor for low-latency action recognition and
  detection,'' in \emph{Proc. Int. Conf. Comput. Vision}, 2013.

\bibitem{koehn2009statistical}
P.~Koehn, \emph{Statistical machine translation}.\hskip 1em plus 0.5em minus
  0.4em\relax Cambridge University Press, 2009.

\bibitem{Hussein2013}
M.~E. Hussein, M.~Torki, M.~A. Gowayyed, and M.~El-Saban, ``Human action
  recognition using a temporal hierarchy of covariance descriptor on 3{D} joint
  locations,'' in \emph{Proc. Int. Joint Conf. Artificial Intelligence}, 2013.

\bibitem{Zhou2014}
L.~Zhou, W.~Li, Y.~Zhang, P.~Ogunbona, D.~T. Nguyen, and H.~Zhang,
  ``Discriminative key pose extraction using extended {LC}-{KSVD} for action
  recognition,'' in \emph{Proc. Int. Conf. Digital lmage Computing Techniques
  Appl.}, 2014.

\end{thebibliography}
}

\section*{Supplementary material:}
In this supplementary material, the textual instructions of the actions in the MSRC-12 gesture~\cite{Fothergill2012} and WorkoutSu-10 exercise~\cite{Negin2013} datasets are described in the first section. The second section details the extracted semantic poses of the two datasets. In the third section, the experimental results on the recognition accuracy vs. the number of visual pose candidates are presented.
\section*{Textual instructions}
\label{text}
Beginning with a starting configuration of the body, the instructions consist of a sequence description of the elementary motions that constitute an action. The instructions can be understood and followed by a subject with average literacy to perform the action. 
\subsection*{MSRC-12}
\begin{itemize}
\item Lift outstretched arms: 
\begin{itemize}
\item Begin in standing position with arms beside body.
\item Raise outstretched arms from sides to should height until overhead.
\item Return to starting position.
\end{itemize}
\item Duck:
\begin{itemize}
\item Begin in standing position with arms beside body.
\item Slightly bend knee.
\item Return to starting position.
\end{itemize}
\item Push right:
\begin{itemize}
\item Begin in standing position with arms beside body.
\item Bring right hand in front of tummy.
\item Slide right hand towards the right until same plane as torso.
\item Return to starting position.
\end{itemize}
\item Goggles:
\begin{itemize}
\item Begin in standing position with arms beside body.
\item Raise hands through tummy, chest until eyes.
\item Return to starting position.
\end{itemize}
\item Wind up:
\begin{itemize}
\item Begin in standing position with arms beside body.
\item Raise arms through tummy, chest until shoulder height.
\item Bring arms back and and return to starting position.
\end{itemize}
\item Shoot:
\begin{itemize}
\item Begin in standing position with arms beside body.
\item Stretch arms out in front of body with clasped hands.
\item Slightly bring up hands.
\item Return to starting position.
\end{itemize}
\item Bow:
\begin{itemize}
\item Begin in standing position with arms beside body.
\item Bend forwards with torso parallel with the floor.
\item Return to starting position.
\end{itemize}
\item Throw:
\begin{itemize}
\item Begin with standing position.
\item Step left foot forward while raising right hand overhead and back to the point of maximum external shoulder rotation.
\item Accelerate right arm forward until elbow is completely straight.
\item Return to starting position.
\end{itemize}
\item Had enough:
\begin{itemize}
\item Begin with standing position.
\item Raise arms with hands on the head.
\item Return to starting position.
\end{itemize}
\item Change weapon:
\begin{itemize}
\item Begin with standing position.
\item Reach over left shoulder with right hand.
\item Bring both hands in front of tummy as if they are holding something.
\item Return to starting position.
\end{itemize}
\item Beat both:
\begin{itemize}
\item Begin with standing position.
\item Raise hands to the head level.
\item Beat hands sidewards for a few times.
\item Return to starting position.
\end{itemize}
\item Kick:
\begin{itemize}
\item Begin with standing position.
\item Step left foot forward while raising right foot to the right knee.
\item Straighten right leg forward.
\item Return to starting position.
\end{itemize}
\end{itemize}
\subsection*{WorkoutSu-10}
\label{workout}
\begin{itemize}
\item Hip flexion ($A_1$)
\begin{itemize}
\item Begin in standing position with arms beside body.
\item Place hands on the waist.
\item Flex left leg at the hip up to 90 degree and bend the knee.
\item Return to starting position.
\end{itemize}
\item Torso rotation ($A_2$)
\begin{itemize}
\item Begin in standing position with arms beside body.
\item Outstretch arms at shoulder height with clasped hands and flex left leg at knee.
\item Rotate torso towards left side at almost 45 degrees.
\item Rotate torso to the right.
\item Return to starting position.
\end{itemize}
\item Lateral stepping ($A_3$)
\begin{itemize}
\item Begin in standing position with arms beside body.
\item Place hands on the waist.
\item With right foot step laterally to the right.
\item Bring left foot beside right foot.
\item Repeat the last two elementary actions for several times.
\item With left foot step laterally to the left.
\item Bring right foot beside left foot.
\item Repeat the last two elementary actions for several times.
\item Return to starting position.
\end{itemize}
\item Thoracic rotation ($B_1$)
\begin{itemize}
\item Begin in standing position with arms are flexed at elbow and hands raised at shoulder height.
\item Rotate torso to left side.
\item Return to starting position.
\end{itemize}
\item Hip Adductor stretch ($B_2$)
\begin{itemize}
\item Begin in standing position with arms beside body.
\item Spread legs beyond shoulder width.
\item Shift body weight to right leg by bending the knee up to 90 degree and straighten left leg.
\item Return to starting position.
\end{itemize}
\item Hip adductor stretch ($B_3$)
\begin{itemize}
\item Begin in standing position with arms beside body.
\item Spread legs beyond shoulder width.
\item Bend forwards with torso parallel with the floor while arms are hanging down.
\item Return to starting position.
\end{itemize}
\item Curl-to-press ($C_1$)
\begin{itemize}
\item Begin in standing position with arms beside body.
\item Flex arms at elbow and raise arms in front of body and overhead.
\item Return to starting position.
\end{itemize}
\item Freestanding squarts ($C_2$)
\begin{itemize}
\item Begin in standing position with arms beside body.
\item Bend knee up to 90 degrees and arms are outstretched at shoulder height in front of the body.
\item Return to starting position.
\end{itemize}
\item Transverse horizontal punch ($C_3$)
\begin{itemize}
\item Begin in standing position with arms beside body.
\item Rotate torso towards left side and punch with right hand in front of the body until arm is straight.
\item Rotate torso towards right side and punch with left hand in front of the body until arm is straight.
\item Return to starting position.
\end{itemize}
\item Oblique stretch ($C_4$)
\begin{itemize}
\item Begin in standing position with arms beside body.
\item Spread legs beyond shoulder width.
\item Raise left arm laterally overhead and lean torso towards the right.
\item Return to starting position.
\end{itemize}
\end{itemize}
\section*{Semantic poses}
\label{semantic}
Semantic poses extracted from the MSRC-12 gesture and WorkoutSu-10 exercise datasets are show in Table~\ref{tab2} and~\ref{tab4} and their linguistic descriptions are as follows. 
\begin{description}
\item [$T_1$] Arms are beside legs. 
\item [$T_2$] Arms are overhead with elbows above shoulder level. 
\item [$T_3$] Thighs at an angle to the shins.
\item [$T_4$] Right arm is in front of stomach. 
\item [$T_5$] Horizontally outstretched right arm is in the right of the body. 
\item [$T_6$] Hands are on the corresponding eyes. 
\item [$T_7$] Two arms are in front of stomach. 
\item [$T_8$] Horizontally outstretched arms are in front of the body. 
\item [$T_9$] Torso is parallel to floor. 
\item [$T_{10}$] Left foot is one step in front of the right foot. Right elbow is at shoulder level. 
\item [$T_{11}$] Left foot is one step front of the right foot. Horizontally outstretched right arm is in front of the body. 
\item [$T_{12}$] Hands are on the head. 
\item [$T_{13}$] Horizontally outstretched right arm is in front of the body.  
\item [$T_{14}$] Two elbows are at shoulder level.  
\item [$T_{15}$] Left foot is one step in front of the right foot.
\item [$T_{16}$] Raised and outstretched right leg is in front of the body.  
\item [$T_{17}$] Forward raised left knee is with thigh at almost 90 degrees to the shin.  
\item [$T_{18}$] The trunk faces the left side at almost 45 degrees. Horizontally outstretched arms are in front of the body. Left shin is at almost same level as right knee.  
\item [$T_{19}$] The trunk faces the right side at almost 45 degrees. Horizontally outstretched arms are in front of the body. Right shin is at almost same level as left knee. 
\item [$T_{20}$] Two hands on the hips. Thighs at a slight angle to the shins. 
\item [$T_{21}$] Raised two arms are beside the body with hands at almost shoulder level.  
\item [$T_{22}$] The trunk faces one side at almost 45 degrees. Raised two arms are beside the body with hands at almost shoulder level. 
\item [$T_{23}$] One thigh at an angle to the corresponding shin. The other leg is outstretched. 
\item [$T_{24}$] Separate legs are much wider than shoulder.
\item [$T_{25}$] Separate legs are much wider than shoulder. Torso is parallel to floor and hanging arms are down besides legs.  
\item [$T_{26}$] Trunk faces the left side at almost 90 degrees. Horizontally outstretched right arm is in front of the body.   
\item [$T_{27}$] Trunk faces the right side at almost 90 degrees. Horizontally outstretched left arm is in front of the body.   
\item [$T_{28}$] The trunk is bent obliquely to one side. The stretched opposite side arm is over the head. 
\end{description}

\begin{table}[!t]
\centering
\caption{Semantic poses used in MSRC-12 dataset.}
\begin{tabular}{|l|l|}
\hline 
Gestures &Semantic poses \\
\hline
Lift outstretched arms &$T_1$, $T_2$ \\
Duck  &$T_1$, $T_3$ \\
Push right   &$T_1$, $T_4$, $T_5$\\
Goggles & $T_1$, $T_6$  \\
Wind up & $T_1$, $T_7$, $T_2$, $T_7$\\
Shoot& $T_1$, $T_8$\\
Bow & $T_1$, $T_{9}$\\
Throw & $T_1$, $T_{10}$, $T_{11}$\\
Had enough  & $T_1$, $T_{12}$\\
Change weapon & $T_1$, $T_{13}$, $T_{7}$\\
Beat both &$T_1$, $T_{14}$, $T_2$\\
Kick&$T_1$, $T_{15}$, $T_{16}$\\
\hline
\end{tabular}

\label{tab2}
\end{table}

\begin{table}[!t]
\centering
\caption{Semantic poses used in WorkoutSu-10 dataset.}
\begin{tabular}{|l|l|}
\hline 
Exercises &Semantic poses \\
\hline
Hip flexion ($A_1$) &$T_1$, $T_{17}$ \\
Trunk rotation ($A_2$) &$T_1$, $T_{18}$, $T_{19}$ \\
Lateral stepping ($A_3$)   &$T_1$, $T_{20}$\\
Thoracic rotation ($B_1$)&$T_{21}$, $T_{22}$  \\
Hip adductor stretch ($B_2$) & $T_1$, $T_{23}$\\
Hip stretch ($B_3$) & $T_1$, $T_{24}$, $T_{25}$\\
Curl-to-press ($C_1$)& $T_1$, $T_{2}$\\
Freestanding squats ($C_2$)& $T_1$, $T_{3}$\\
Transverse horizontal punch ($C_3$)& $T_1$, $T_{26}$, $T_{27}$\\
Oblique stretch ($C_4$) & $T_1$, $T_{28}$\\
 \hline
\end{tabular}

\label{tab4}
\end{table}
\section*{Impact of the number of visual pose candidates}
\label{moretest}

\begin{figure}[!t]
\begin{center}
   \includegraphics[width=0.85\linewidth]{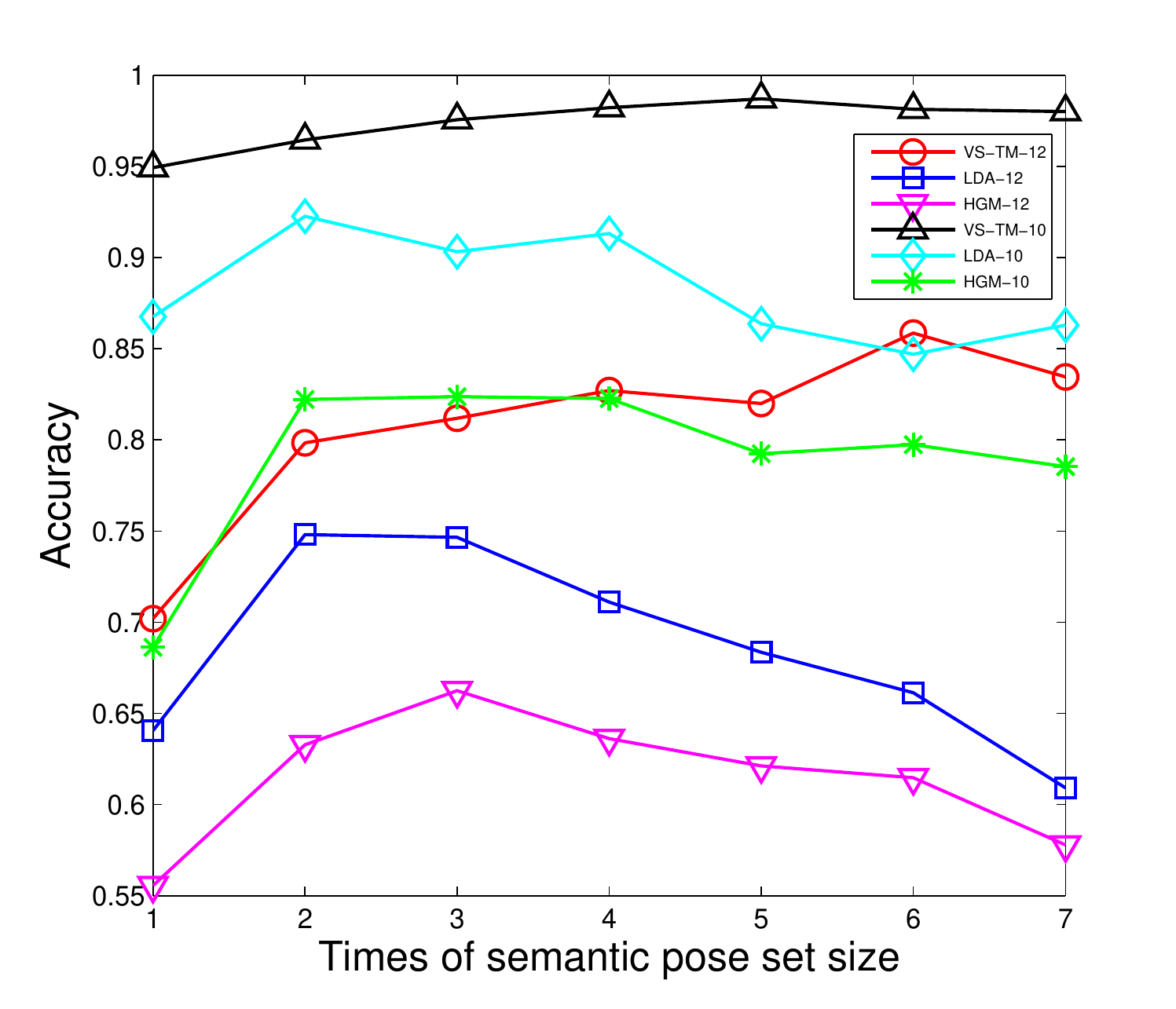}
\end{center}
   \caption{Recognition accuracy versus the number of visual pose candidates.}
\label{fig_relation}
\end{figure}
The recognition accuracy versus the number of visual pose candidates is shown in Figure~\ref{fig_relation}, where 12 and 10 in the legends represent MSRC-12 and WorkoutSu-10 datasets respectively.  Specifically, the number of visual pose candidates ranges from one to seven times as many as the number of semantic poses. Results have shows that the recognition rate of the propose method increases first with the increase of the number of visual pose candidates and then levels off when the number of visual pose candidates reaches 5 or 6 times as many as the number of semantic poses. 

We also compare the proposed method with the latent Dirichlet allocation (LDA) model~\cite{Niebles2008} and a hierarchical generative model 
(HGM)~\cite{yang2014hierarchical} on the MSRC-12 and WorkoutSu-10 datasets, where the number of words was set to one to seven times as many as the number of topics in LDA and the number of global topics in HGM respectively. Experimental results have shown that the proposed method is more robust than LDA and HGM with respect to the number of visual pose candidates.

\end{document}